%% file: RoboSeed.tex
\documentclass{article}
\usepackage{roboseed_report}

\usepackage{fontspec}
\usepackage{booktabs}
\usepackage{caption}
\usepackage{enumitem}
\usepackage{titlesec}
\usepackage{tocloft}
\usepackage{tabularx}
\usepackage{longtable}
\usepackage{array}
\usepackage{pdflscape}
\usepackage{float}

\renewcommand{\arraystretch}{1.16}

\newcolumntype{Y}{>{\centering\arraybackslash\hyphenpenalty=10000\exhyphenpenalty=10000}X}
\newcolumntype{C}[1]{>{\centering\arraybackslash\hyphenpenalty=10000\exhyphenpenalty=10000}m{#1}}

\titleformat*{\section}{\large\bfseries\color{darkblue}}
\titleformat*{\subsection}{\normalsize\bfseries\color{darkblue}}
\titleformat*{\subsubsection}{\normalsize\bfseries\color{darkblue}}

\definecolor{contentsblue}{HTML}{0066CC}
\hypersetup{linkcolor=contentsblue,citecolor=contentsblue,urlcolor=contentsblue,linktoc=section}

\cftsetindents{subsubsection}{2.3em}{3.6em}

\usepackage{amsmath,amssymb}
\usepackage{multirow}
\setlist{nosep,leftmargin=1.8em}
\newcommand{\fig}[4]{\begin{figure}[!htbp]\centering\includegraphics[width=#2\linewidth]{figures/#1.pdf}\caption{#3}\label{#4}\end{figure}}


\title{RoboSeed: Multidimensional Evaluation of\\Vision--Language--Action and World--Action Models\\[0.35em]\large Capability, Resource Affordability, and Research Evidence}
\input{sections/authors.tex}
\date{October 2026}
\begin{document}
\setlength{\headheight}{38pt}
\maketitle
\begin{center}\small\textbf{Project repository:} \url{https://github.com/PrimitiveX/RoboSeed}\end{center}
\begin{abstract}
Choosing a robot action model requires evidence about task capability, adaptation cost, deployment requirements, and the reliability of reported conclusions. Success rates alone offer limited guidance to researchers with different computing budgets and goals. We propose RoboSeed, a multidimensional evaluation framework for vision-language-action (VLA) and world-action models (WAM), within a continuing program to train, evaluate, and openly release robot models with verifiable evidence for model selection.

The framework defines five dimensions: capability, downstream training affordability, deployment affordability, locally verified research value, and reproduction evidence. It maps empirical measurements and audit criteria to 0-100 profiles using fixed normalization anchors, with balanced, training-constrained, and deployment-constrained preference settings. Interface and protocol checks establish when comparisons are appropriate. Missing measurements remain explicit, incomplete aggregate scores are withheld, and weighted input coverage identifies evidence gaps.

We apply the framework to representative VLA and WAM models, drawing on local training, released-weight evaluation, task-level audits, and an evaluation-system study. The results reveal budget-dependent backbone choices, sensitivity to adaptation recipes, and differences between language and visual task interfaces. Reproducible scoring code and machine-readable evidence support extension to additional models. The study provides dimension-level assessments and priorities for further measurement, while incomplete serving traces and heterogeneous protocols preclude overall numerical rankings.

\end{abstract}
\vspace{0.6em}
\roboseedauthornotes
\clearpage
\setlength{\headheight}{24pt}
\setcounter{tocdepth}{3}\tableofcontents\clearpage

\section{Introduction}
\subsection{From Task Success to Research Decisions}
Community researchers choosing a robot action model face several linked questions: can it solve the target tasks, can the team afford to adapt and deploy it, and does the available evidence justify investing in its proposed mechanism? VLA models and WAMs offer different combinations of semantic conditioning, predictive representations and action generation. Their practical value depends on the task interface, computing budget and strength of the experimental evidence as well as the final success rate.

A reported percentage leaves much of this decision unresolved. Two checkpoints with similar success can require different adaptation budgets, observation pipelines and serving resources. A strong released checkpoint can coexist with a fragile local training recipe. A compact policy may rely on task-specific visual conditions rather than accept arbitrary language instructions. Researchers therefore need an evaluation that connects task outcomes to resource requirements and the conclusions supported by the experiment.

\textbf{Our goal is to continuously train, evaluate, and openly release robot models for different computing budgets, while providing verifiable evidence for model selection.} The report, model checkpoints, experimental results, and reproduction materials are intended to form a connected resource that community researchers can inspect, use, and extend. We plan to publish our trained checkpoints and associated experimental results on Hugging Face, with explicit links to the corresponding configurations and report versions. Public checkpoint download entries remain pending in the current snapshot.

RoboSeed addresses the model-selection need by proposing a \textbf{multidimensional evaluation and scoring framework} for the research community. We organize model selection around a common set of criteria, make their scoring rules reproducible, and connect each judgment to its underlying measurements. Training and evaluation across VLA and WAM families supply the empirical basis for these assessments. This organization lets researchers use the same evaluation questions across different architectures while preserving native task interfaces and experimental conditions.

\subsection{The Proposed RoboSeed Evaluation Framework}
Our framework represents each candidate by five dimensions on a 0--100 scale: capability $P$, downstream training affordability $T$, deployment affordability $D$, locally verified research value $V$, and reproduction evidence $R$. Table~\ref{tab:framework_overview} connects these dimensions to the decisions they support. The evaluation unit includes the checkpoint, training recipe, execution interface and evaluator, so a score remains tied to a concrete experimental setting.

\begin{table}[!htbp]\centering\small
\caption{The proposed evaluation framework connects five assessment dimensions to community research decisions. Chapter~\ref{sec:assessment} defines the scoring rules and reports available scores; deployment measurements remain incomplete in this snapshot.}\label{tab:framework_overview}
\begin{tabularx}{\linewidth}{C{3.5cm}YY}\toprule
Dimension & Evaluation basis & Research decision\\\midrule
Capability ($P$) & Clean and Randomized closed-loop task outcomes & Does the candidate meet the required task-success level?\\\addlinespace[3pt]
Training affordability ($T$) & Measured downstream H200 GPU-hours & Can the team afford adaptation and repeated experiments?\\\addlinespace[3pt]
Deployment affordability ($D$) & Weight size, inference peak memory and p95 latency & Can the policy fit the serving budget and response requirement?\\\addlinespace[3pt]
Research value ($V$) & Inspected mechanism, comparators, controlled ablations and transfer evidence & Which conclusions can inform further mechanism research?\\\addlinespace[3pt]
Reproduction evidence ($R$) & Configuration mapping, task records, coverage, training records and seeds & How well can the reported result be checked and reproduced?\\\bottomrule
\end{tabularx}\end{table}

Three design choices make the profiles reusable. First, fixed normalization anchors give cost scores a declared budget interpretation and keep existing scores stable as candidates are added. Second, balanced, training-constrained and deployment-constrained weight settings expose how research priorities affect aggregate preferences. Third, explicit missing-data rules retain unavailable measurements as dashes and withhold incomplete composites; weighted input coverage identifies which observations remain to be collected. Raw metrics accompany the scores so researchers can change anchors and weights for their own budgets.

The intended use is to check task and execution compatibility, enforce capability and resource requirements, and then inspect the feasible candidates' dimension profiles under declared preferences. The research-value and reproduction-evidence dimensions further identify useful hypotheses and gaps in experimental support. This provides a broader basis for community evaluation while keeping the interpretation of each dimension explicit. The weights and thresholds are a versioned decision policy; their usefulness across additional laboratories and applications remains to be tested.

\subsection{Contributions}
\textbf{A multidimensional evaluation framework for community model selection.} We propose five complementary scoring dimensions that bring capability, training and deployment affordability, locally established research value, and reproduction evidence into one assessment. Fixed scoring anchors, explicit audit criteria, alternative preference profiles and missing-data rules make the framework inspectable and adaptable to different research budgets. Chapter~\ref{sec:assessment} specifies the complete method.

\textbf{An empirical assessment across VLA and WAM families.} We apply the framework to 12 checkpoint units from four policy families, distinguishing released-weight evaluation, local adaptation and unavailable measurements. Backbone comparisons, within-family recipe comparisons and task-level diagnosis connect the scores to concrete decisions: selecting an affordable configuration, identifying a recipe gap, or determining which experiment is needed before attributing a gain to a mechanism. Native conditioning, inherited pre-train and evaluator differences remain attached to every comparison.

\textbf{An auditable basis for model reuse and continuing evaluation.} We provide scoring code, machine-readable profiles, source manifests, training observations and task-level outcomes, alongside a systems study of evaluation throughput. These artifacts connect model-selection conclusions to the experiments that support them. Planned checkpoint releases will pair trained weights with effective configurations, normalization statistics, compatible implementation revisions and corresponding evaluation results. As models and evidence are added, versioned report updates will retain the provenance and completion status of each experimental unit, allowing researchers to reuse the models and reassess their choices.

\fig{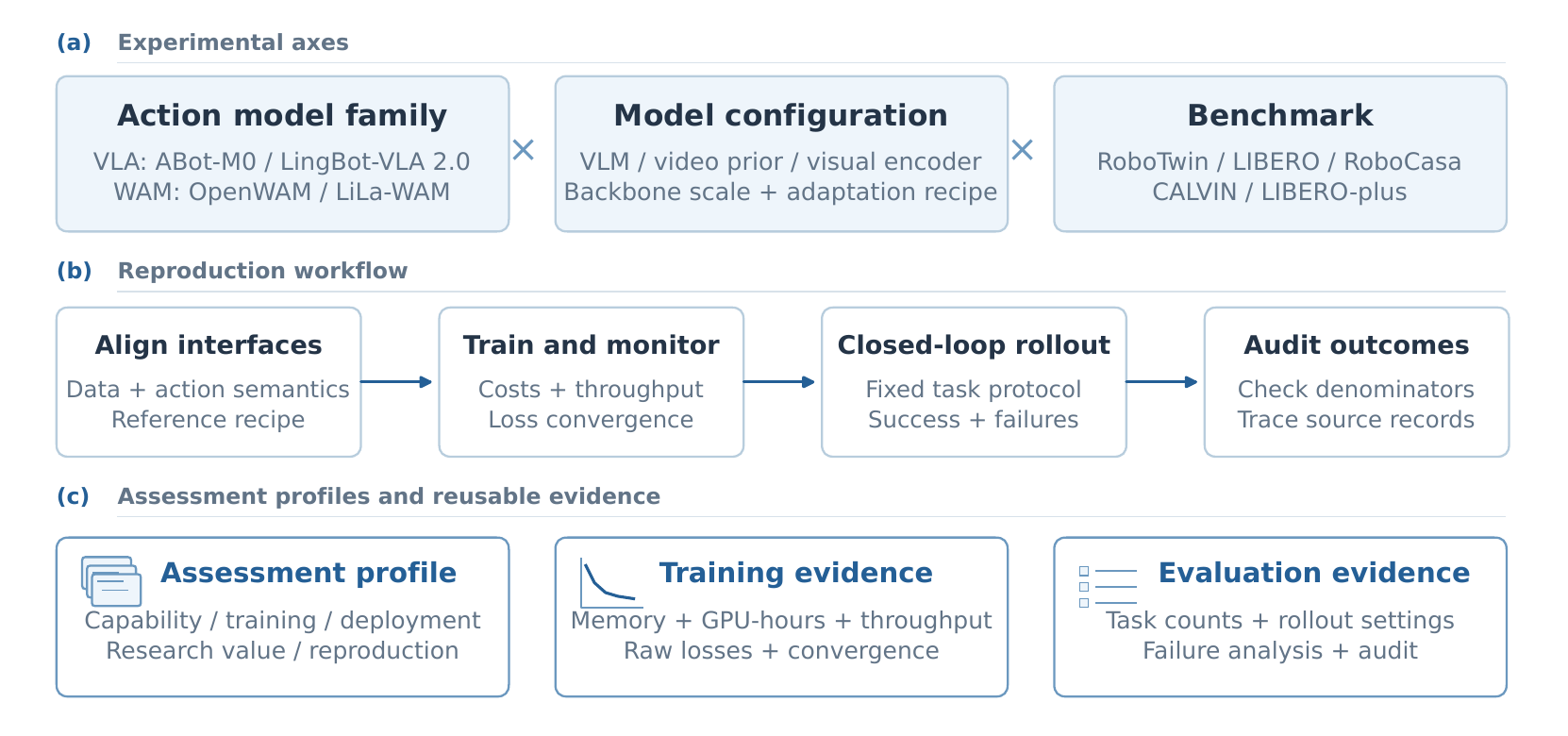}{1}{The RoboSeed evaluation program. (a) Four VLA/WAM families, their native model configurations and benchmarks supply the empirical assessment matrix. (b) Interface alignment, training measurements, closed-loop outcomes and evidence auditing produce the inputs to the proposed framework. (c) Five-dimensional assessment profiles are accompanied by reusable training and evaluation evidence. Chapter~\ref{sec:assessment} defines scoring anchors, preference weights and missing-data rules; Chapters~\ref{sec:vla} and~\ref{sec:wam} provide the supporting model studies. The October 28 audit includes completed ABot Qwen3.5-2B and local LiLa evaluations, with the earlier evidence retained where unchanged.}{fig:project}

\subsection{Empirical Scope and Report Organization}
The October 28 evidence refresh covers six complete ABot-M0 backbone-scale comparisons on RoboTwin and Qwen3.5-4B experiments on LIBERO, LIBERO-plus, RoboCasa GR1 Tabletop, and CALVIN. LIBERO, RoboCasa, and CALVIN use independently trained, dataset-specific checkpoints; LIBERO-plus evaluates the LIBERO checkpoint without additional training. The previously omitted Qwen3.5-2B evaluation is now verified over 10,000 episodes. LingBot-VLA 2.0 and OpenWAM retain their official/local results; LiLa-WAM now includes both official and local stage-2 Clean evaluation. New PX runs extend systems evaluation to all 50 Randomized tasks. An OpenWAM Wan2.1-VACE-1.3B training prefix is documented separately without assigning a success score. Source hashes, incomplete runs and access limitations are recorded in the audit materials.

The report follows the path from empirical evidence to assessment. Chapter~\ref{sec:vla} studies VLA capability, backbone costs and adaptation sensitivity through ABot-M0 and LingBot-VLA 2.0. Chapter~\ref{sec:wam} studies predictive action models, inherited priors and task-interface requirements through OpenWAM and LiLa-WAM. Chapter~\ref{sec:assessment} formalizes our proposed five-dimensional scoring framework, reports all 12 checkpoint profiles, and develops budget-specific choices and measurement priorities. Section~\ref{sec:harness} then examines evaluation-system efficiency. Our contributions comprise the evaluation framework, its empirical application and its auditable implementation; the studied action architectures remain attributed to their original authors. All newly audited model outcomes are simulated RoboTwin evaluations.

\section{Related Work}
\textbf{Open and compact VLA models.} OpenVLA releases a 7B VLA model and fine-tuning tools for reproducing and adapting generalist policies\citep{openvla}. SmolVLA emphasizes affordable training, compact models, and efficient execution\citep{smolvla}. These projects motivate evaluating capability together with resource requirements. Previous work also examines model scale and efficiency; RoboSeed contributes a common assessment framework that connects such capability and cost observations to explicit research-value and reproduction-evidence criteria, supported by an expanding experimental matrix.

\textbf{Action modeling and spatial priors.} ABot-M0 combines VLM semantic features, an action expert, and optional spatial modules, and introduces Action Manifold Learning (AML)\citep{abot}. Its modular structure supports backbone substitution. Our experiments retain the main action architecture and training protocol while assessing different backbones under matched task settings.

\textbf{Predictive representations and world--action models.} LingBot-VLA 2.0 uses predictive visual supervision and sparse action experts\citep{lingbot2}; OpenWAM studies how dedicated world and action capacity interact\citep{openwam}; LiLa-WAM uses a compact latent prediction space and language-free visual task conditions\citep{lila}. Their distinct inherited priors, observation interfaces and task specifications motivate evaluation beyond a single success percentage. Our local evidence is assessed independently of the original papers' broader claims.

\textbf{Closed-loop evaluation and robustness.} RoboTwin 2.0 provides bimanual manipulation tasks with domain randomization\citep{robotwin}. LIBERO supports evaluation of spatial, object, goal, and long-horizon manipulation\citep{libero}, while LIBERO-plus probes robustness across seven perturbation dimensions\citep{liberoplus}. RoboCasa GR1 Tabletop includes arms, dexterous hands, and waist control\citep{robocasa}. These environments test different capabilities; their task distributions, action spaces, and evaluation denominators must remain explicit when interpreting results.

\textbf{Continuous multi-task evaluation.} CALVIN provides language-conditioned manipulation in a simulated tabletop environment\citep{calvin}. Its long-horizon evaluation measures the successful prefix of a chain of five instructions. This complements isolated task success: later instructions are attempted only after earlier instructions succeed, and the physical state is carried forward within a chain. We report this composition metric separately from the episode success rates of the other benchmarks.

\section{Empirical Evaluation of Vision--Language--Action Models}\label{sec:vla}
\subsection{Evaluation Protocol and Evidence Collection}\label{sec:reproduction}
This chapter supplies the VLA evidence for the proposed evaluation framework. ABot-M0 backbone experiments quantify capability and downstream training-cost trade-offs, while LingBot-VLA 2.0 official/local comparisons examine adaptation sensitivity. Architecture and interface inspection support the research-value audit; configurations, training records and task-level outcomes support the reproduction-evidence audit. Deployment observations are retained in their native units until they meet the measurement requirements in Chapter~\ref{sec:assessment}. Each model's subsection connects its mechanism, experimental settings and verified results to these evaluation questions.

The shared workflow begins with the source implementation and a documented reference recipe. It checks dataset and action semantics, reconstructs the required interfaces, runs training and closed-loop evaluation, and records both capability and resource costs. Reproduction of a published setting and deliberate changes to that setting are identified explicitly, so a backbone substitution or a run without the robotic pre-train stage is interpreted in its actual context. The current report includes four policy families and a separate systems study. A released checkpoint's successful rollout is distinguished from reproducing its training procedure or isolating its proposed mechanism.

\subsubsection{Experimental Axes and Comparison Principles}
Let $\mathcal{B}$ denote VLM backbones and scales, $\mathcal{D}$ training datasets, and $\mathcal{M}$ action methods. The experimental space is
\begin{equation}
\mathcal{E}=\mathcal{B}\times\mathcal{D}\times\mathcal{M}.
\end{equation}
Each experimental unit also records its training protocol, random seed, execution interface, and evaluator version. The three sets need not have equal cardinalities. Coverage expands as datasets, methods, and backbones are added, with the completion status of each unit recorded explicitly.

Backbone comparisons prioritize matching training data, update count, global batch size, optimizer, action-expert type, frozen modules, and evaluation protocol. Cross-attention and spatial-projection interfaces are adapted to each backbone's hidden width. \textbf{Equal update counts do not imply equal compute}: backbone size, interface width, and kernel implementations affect per-step costs. We therefore report task performance together with observed resource consumption rather than interpreting the study as a strictly equal-FLOPs scaling law.

\subsubsection{Artifacts for Reproducible Releases}
A RoboSeed release is intended to contain a complete experimental unit. Table~\ref{tab:protocol} describes the proposed artifact protocol; Appendix~\ref{app:details} distinguishes available evidence from missing measurements.
\begin{table}[!htbp]\centering\small
\caption{RoboSeed artifact protocol. Each release should state its evidence completeness.}\label{tab:protocol}
\begin{tabularx}{\linewidth}{C{2.8cm}YY}\toprule
Artifact & Contents & Purpose\\\midrule
Configuration and environment & Dataset versions, backbone sources, code commits, dependencies, seeds, and effective configuration snapshots & Reconstruct execution conditions\\\addlinespace[3pt]
Training process & Updates, batch size, learning rates, frozen modules, loss curves, memory, throughput, wall time, and GPU-hours & Assess budgets and convergence\\\addlinespace[3pt]
Model weights & Checkpoints, matching configurations, normalization statistics, file sizes, and checksums & Keep weights and execution interfaces consistent\\\addlinespace[3pt]
Rollout process & Scenes, instructions, episodes, action chunks, step limits, concurrency, runtime, retries, and exclusion rules & Verify denominators and execution settings\\\addlinespace[3pt]
Results and diagnosis & Task-level counts, failure videos, anomalies, diagnostic evidence, and untested hypotheses & Support verification and improvement\\\bottomrule
\end{tabularx}\end{table}

\subsubsection{Metrics and Aggregation}
For task $t$, success rate is $\mathrm{SR}_t=s_t/n_t$. The task macro-average is $T^{-1}\sum_t\mathrm{SR}_t$, and the episode micro-average is $\sum_t s_t/\sum_t n_t$. Each RoboTwin scene evaluates every task for 100 episodes, so these averages coincide within a scene. The two-scene average is $(\mathrm{SR}_{\mathrm{Clean}}+\mathrm{SR}_{\mathrm{Randomized}})/2$. LIBERO and RoboCasa likewise use equal episode counts per task.

LIBERO-plus evaluates each perturbation variant once, with different numbers of variants across suites and categories. Its \textbf{overall score is the micro-average across all variants}; category scores use their actual denominators. Scores from different benchmarks are not pooled. Differences between standard LIBERO and LIBERO-plus describe performance under their respective test distributions.

For CALVIN, let $\ell_i\in\{0,\ldots,5\}$ be the successful prefix length of sequence $i$, and $C_k=N^{-1}\sum_i\mathbf{1}[\ell_i\geq k]$. The primary metric is
\begin{equation}
\overline{\ell}=\frac{1}{N}\sum_{i=1}^N\ell_i=\sum_{k=1}^5 C_k.
\end{equation}
We also report $C_1,\ldots,C_5$ and task-specific successes divided by actual attempts. Task denominators are unequal because a failed subtask terminates its chain; these conditional rates are diagnostic and are not substituted for the sequence metric.

\subsection{ABot-M0}\label{sec:abot}
\subsubsection{Architecture and Backbone Substitution}
Figure~\ref{fig:architecture} shows the implemented module relationships. Multi-view images and language instructions are encoded by the VLM. A VGGT stream provides geometric features, which are projected to the backbone's hidden width and fused through cross-attention. The action expert conditions on fused features, available robot state, and noisy actions to predict an action sequence. Predictions are unnormalized using dataset statistics and passed to the controller.

\fig{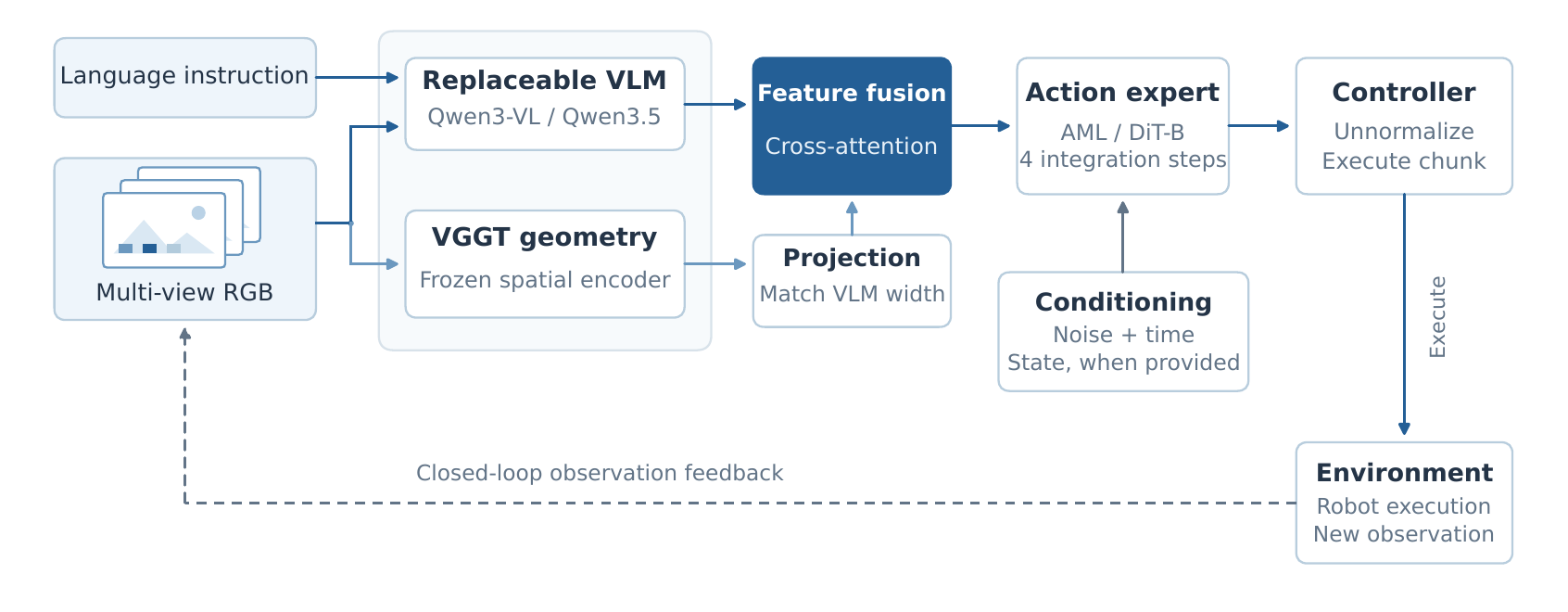}{1}{ABot-M0 backbone substitution and action generation, redrawn from the implementation interfaces. Feature widths are adapted when replacing the VLM; VGGT is frozen during these experiments. Dark blue arrows denote semantic and action flow, light blue arrows denote geometry flow, and the dashed connection denotes observation feedback from execution. State conditioning is used when supplied by the execution interface.}{fig:architecture}

\subsubsection{Action Learning Objective}
For an expert action chunk $A$, noise $\epsilon$, and time $\tau\in[0,1)$, the noisy input is $A^\tau=\tau A+(1-\tau)\epsilon$. The action expert predicts a clean chunk $\hat A=f_\theta(\phi,q,A^\tau,\tau)$, where $\phi$ denotes fused features and $q$ the state condition when supplied. The implementation derives velocity fields from the predicted and target action chunks:
\begin{equation}
\hat v=\frac{\hat A-A^\tau}{\max(1-\tau,\delta)},\quad
v=\frac{A-A^\tau}{\max(1-\tau,\delta)},\quad
\mathcal{L}=\mathbb{E}\!\left[\|\hat v-v\|^2\right].
\end{equation}
Here $\delta$ ensures numerical stability. When action dimensions are padded, the loss is computed over valid dimensions. All loss curves in this report come from actual training logs. They describe optimization of the learning objective; task execution is assessed by closed-loop success.

\subsubsection{Experimental Setup}
\begin{table}[!htbp]\centering\small
\caption{Verified experimental coverage in the current snapshot. Complete scores are derived from original evaluation artifacts.}\label{tab:coverage}
\begin{tabularx}{\linewidth}{C{2.35cm}C{4.35cm}C{3.05cm}Y}\toprule
Benchmark & Backbone & Training protocol & Evaluation coverage\\\midrule
RoboTwin & \shortstack{Qwen3.5-0.8B / 2B / 4B / 9B\\Qwen3-VL-2B / 4B} & \shortstack{150k updates\\Final weights} & 50 tasks $\times$ 100 episodes per scene; 2 scenes\\\addlinespace[3pt]
LIBERO & Qwen3.5-4B & 40k checkpoint & 4 suites; 10 tasks per suite; 50 episodes per task\\\addlinespace[3pt]
LIBERO-plus & \shortstack{LIBERO-trained\\Qwen3.5-4B} & \shortstack{Zero-shot\\No extra training} & 10,030 variants, 1 episode each\\\addlinespace[3pt]
RoboCasa GR1 & Qwen3.5-4B & 50k checkpoint & 24 tasks $\times$ 50 episodes\\\addlinespace[3pt]
CALVIN & Qwen3.5-4B & \shortstack{34k updates\\Final weights} & ABCD $\rightarrow$ D; 1,000 sequences of up to 5 tasks\\\bottomrule
\end{tabularx}\end{table}

\textbf{RoboTwin training.} All six evaluated configurations use four H200 NVL GPUs, global batch size 64 (16 per GPU, accumulation 1), and 150,000 updates. They initialize from the corresponding VLM and proceed directly to the downstream post-train stage (supervised fine-tuning, SFT), without the unified robotic pre-train stage. These scores therefore describe the reproduced training setting rather than the official pre-trained release. The training data include Clean and Randomized demonstrations; experimental records report 27,500 trajectories and 6,075,103 transitions. Image resolution is $224\times224$, the training seed is 42, and the spatial module is frozen. Base and VLM learning rates are $10^{-5}$; the action-expert rate is $10^{-4}$. AdamW uses $\beta=(0.9,0.95)$, with 5,000 warmup updates and a cosine schedule to $5\times10^{-7}$. The action horizon is 50; configured state and action dimensions are both 14. The action expert is DiT-B, with four repeated time samples per training example. Dependency details appear in the appendix.

\textbf{RoboTwin evaluation.} Every configuration evaluates 50 tasks for 100 episodes each under both \texttt{demo\_clean} and \texttt{demo\_randomized}, using unseen instructions and launch seed 0. The recorded evaluator commit is \texttt{13c3c47}, with the interface and frame-sampling modifications used during these runs. Scores use binary environment success counts, without mixing in progress scores. Since training includes randomized demonstrations, Randomized scores describe performance in randomized scenes; they do not establish generalization from exclusively Clean training. Comparisons with public leaderboards require matching the evaluator version and execution protocol.

\textbf{Training on other datasets.} LIBERO and RoboCasa are trained independently. LIBERO uses 40k updates, global batch 64, and base/VLM/action learning rates of $2.5\times10^{-5}$, $3\times10^{-5}$, and $10^{-4}$. Its action horizon is 10, configured state dimension 7, and internal action representation 14-dimensional, with the active Franka channels used for execution. The saved training script specifies \texttt{include\_state=false}; its effective behavior requires checking the corresponding implementation version. RoboCasa uses 24 datasets with 1,000 demonstrations each, 50k updates, and learning rates of $10^{-5}$, $10^{-5}$, and $10^{-4}$. Its horizon is 16, state encoding uses sine and cosine of 29 joint values (58 dimensions), and actions are 29-dimensional absolute joint targets. Both experiments freeze the spatial module.

\subsubsection{Backbone Scale and Task Success}
\begin{table}[!htbp]\centering\small
\caption{ABot-M0 backbone comparison on RoboTwin. Rates are percentages; Average equally weights the two scenes.}\label{tab:robotwin}
\setlength{\tabcolsep}{4pt}
\begin{tabular}{cccccc}\toprule
Backbone & Clean count & Clean & Randomized count & Randomized & Average\\\midrule
Qwen3.5-0.8B & 4038/5000 & 80.76 & 4025/5000 & 80.50 & 80.63 \\
Qwen3-VL-2B & 4183/5000 & 83.66 & 4195/5000 & 83.90 & 83.78 \\
Qwen3-VL-4B & 4285/5000 & 85.70 & 4267/5000 & 85.34 & 85.52 \\
Qwen3.5-2B & 4215/5000 & 84.30 & 4197/5000 & 83.94 & 84.12 \\
Qwen3.5-4B & 4294/5000 & 85.88 & 4256/5000 & 85.12 & 85.50 \\
Qwen3.5-9B & 4199/5000 & 83.98 & 4184/5000 & 83.68 & 83.83 \\
\bottomrule\end{tabular}\end{table}

\fig{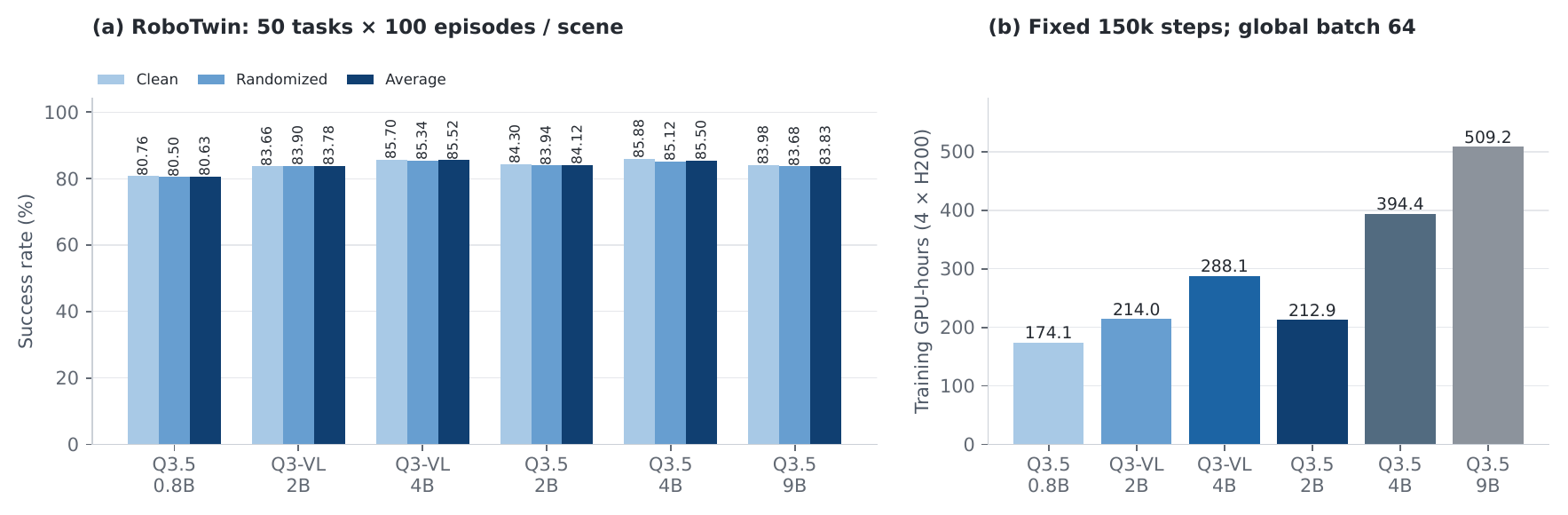}{1}{Task performance and training cost. Every configuration uses 150k updates and global batch 64. Bars start at zero. Training GPU-hours equal four GPUs multiplied by training wall time; they exclude evaluation and do not indicate equal-FLOPs budgets.}{fig:scale}

\textbf{Smaller backbones retain substantial capability.} Qwen3-VL-2B reaches 83.78\%, trailing the same-family 4B score of 85.52\% by 1.74 percentage points, while reducing GPU-hours from 288.1 to 214.0 (approximately 25.7\%). Qwen3.5-0.8B trails its 4B counterpart by 4.87 points while reducing GPU-hours from 394.4 to 174.1 (approximately 55.8\%). This evidence supports smaller backbones as useful options for constrained budgets under the tested recipe. It does not establish that these full VLA configurations can be trained on arbitrary consumer GPUs.

\textbf{The completed Qwen3.5-2B run adds a budget-relevant option.} Its 100 result files match the final 100-episode counters for all 50 tasks in both scenes: 4,215/5,000 Clean (84.30\%) and 4,197/5,000 Randomized (83.94\%), averaging 84.12\%. Recorded training is 212.9 H200 GPU-hours. This point estimate exceeds Qwen3-VL-2B by 0.34 percentage points at a very similar training cost; the single-seed experiment does not establish a robust superiority claim.

\textbf{The two 4B backbones have similar aggregate scores.} Qwen3-VL-4B and Qwen3.5-4B average 85.52\% and 85.50\%. Relative to Qwen3-VL, Qwen3.5 changes Clean success by $+0.18$ percentage points and Randomized success by $-0.22$ points. The experiment provides no clear aggregate success-rate advantage for the newer backbone. A single training seed does not establish statistical equivalence, and differences in architecture or pre-train recipes between families cannot be attributed to parameter count alone.

\textbf{Scaling beyond 4B does not improve this recipe.} The completed Qwen3.5-9B evaluation yields 4,199/5,000 Clean and 4,184/5,000 Randomized successes, for a two-scene average of 83.83\%. This is 1.67 percentage points below Qwen3.5-4B while consuming 509.2 rather than 394.4 training GPU-hours, an increase of approximately 29.1\%. Its average is only 0.05 points above Qwen3-VL-2B, despite approximately 2.38 times the training-device time. Under the tested recipe, expanding from smaller backbones to 4B improves the observed averages, but expanding further to 9B does not. With one training seed per configuration, these observations do not establish a universal optimum or identify the cause of the 9B result.

\textbf{A second 2B family is being evaluated.} Qwen3.5-2B has completed 150k updates and saved final weights. It requires 53.23 hours on four H200 GPUs (212.9 GPU-hours), processes approximately 50.1 nominal samples/s, and saves an 8.33 GB checkpoint. Its final logged action loss is approximately $1.44\times10^{-3}$. RoboTwin evaluation is still in progress at the snapshot cutoff; training completion and low loss are insufficient to assign a task-success score or rank it against Qwen3-VL-2B.

\subsubsection{Training Memory, Runtime, Throughput, and Weights}
\begin{table}[!htbp]\centering\footnotesize
\caption{RoboTwin training process. Approximate memory readings retain the original records' GiB labels; uniformly sampled memory traces are unavailable. Throughput is recomputed from updates, batch size, and complete wall time.}\label{tab:train}
\setlength{\tabcolsep}{4pt}
\begin{tabular}{ccccccc}\toprule
& & & & \multicolumn{2}{c}{Memory / GPU (GiB)} &\\
Backbone & Hours & GPU-h & Samples/s & Mean & Recorded max. & Weights (GB)\\\midrule
Qwen3.5-0.8B & 43.53 & 174.1 & 61.3 & $\sim$52.27 & $\sim$55.06 & 5.04 \\
Qwen3-VL-2B & 53.51 & 214.0 & 49.8 & $\sim$54.90 & $\sim$55.30 & 7.77 \\
Qwen3-VL-4B & 72.03 & 288.1 & 37.0 & $\sim$83.26 & $\sim$83.93 & 12.58 \\
Qwen3.5-2B & 53.23 & 212.9 & 50.1 & $\sim$58.95 & $\sim$71.01 & 8.33 \\
Qwen3.5-4B & 98.59 & 394.4 & 27.0 & $\sim$142.50 & $\sim$143.00 & 13.28 \\
Qwen3.5-9B & 127.30 & 509.2 & 20.9 & $\sim$137.04 & $\sim$139.69 & 21.88 \\
\bottomrule\end{tabular}\end{table}

Training durations for the six evaluated backbones are 43:31:53, 53:30:44, 72:01:59, 53:13:34, 98:35:29, and 127:18:06 in table order. Qwen3.5-2B memory records are now available: approximately 58.95 GiB mean and 71.01 GiB maximum under the source monitoring convention. Throughput is the nominal number of batch samples processed per second, not unique data items. Checkpoint sizes come from actual file byte counts and use decimal GB ($10^9$ bytes). Saved weights include action, spatial, and other modules and cannot be inferred from the nominal VLM size. GPU-hours measure allocated training-device time, rather than energy, utilization, or monetary cost.

\fig{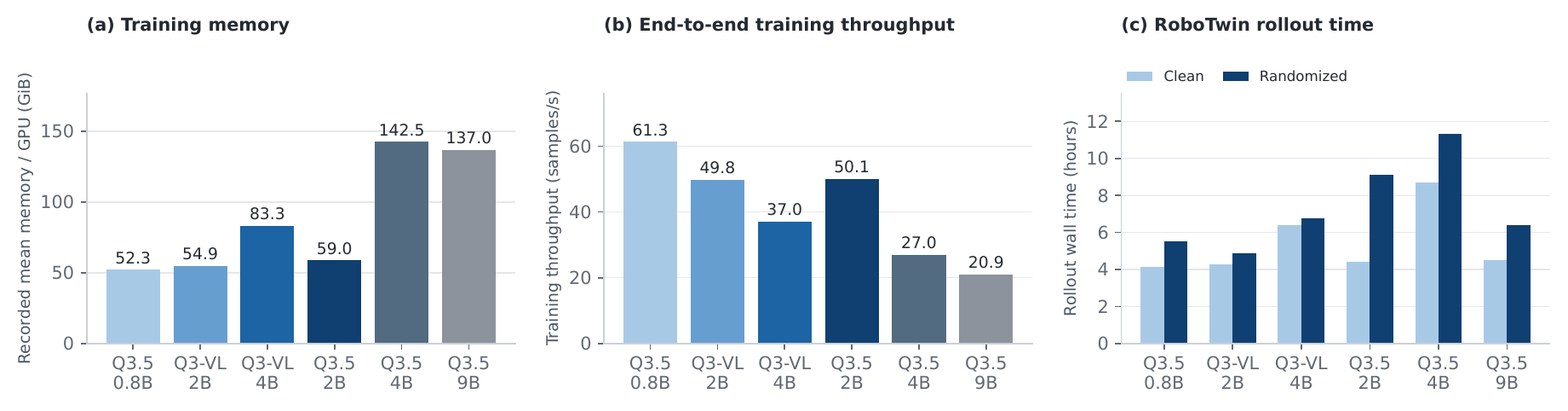}{1}{Training resources and rollout details. Memory bars use approximate recorded readings. The high Qwen3.5-4B reading is associated with its runtime implementation. Evaluation times retain each run's actual concurrency settings and are not measurements of single-policy inference latency.}{fig:resources}

\textbf{Engineering paths matter when comparing costs.} At 4B, Qwen3.5 takes approximately 36.9\% more training wall time and has approximately 26.9\% lower nominal throughput than Qwen3-VL. The troubleshooting records identify unavailable linear-attention acceleration components and a slower PyTorch fallback path. These measurements describe the software environment used in this experiment. The benefits of enabling the accelerated implementation require a new measurement.

Memory evidence combines approximate monitoring readings with the units stated in the original records. In particular, the 142--143 readings need reconciliation with device capacity, monitoring tools, and definitions of allocated and reserved memory before formal release. We preserve these records transparently rather than treating them as exact minimum hardware requirements obtained with a uniform profiler.

The 9B run records approximately 137.04 GiB mean and 139.69 GiB peak memory, a 21.88 GB checkpoint, and 20.9 nominal samples/s. Its recorded memory is below the 4B Qwen3.5 reading despite its larger backbone. Experiment notes report acceleration dependencies installed for later runs, but no matched before/after 4B experiment isolates their effect. Consequently, memory ordering should not be interpreted as a parameter-scaling rule or a measured kernel speedup.

\subsubsection{Loss Convergence and Closed-Loop Monitoring}
Figure~\ref{fig:loss} compares the six backbones through four complementary views: full training trajectories, a shared late-training zoom, last-window statistics, and the relationship between training loss and closed-loop success. Overlaying the curves exposes their similar overall shapes, while the late-stage and outcome views make their differences interpretable.

\fig{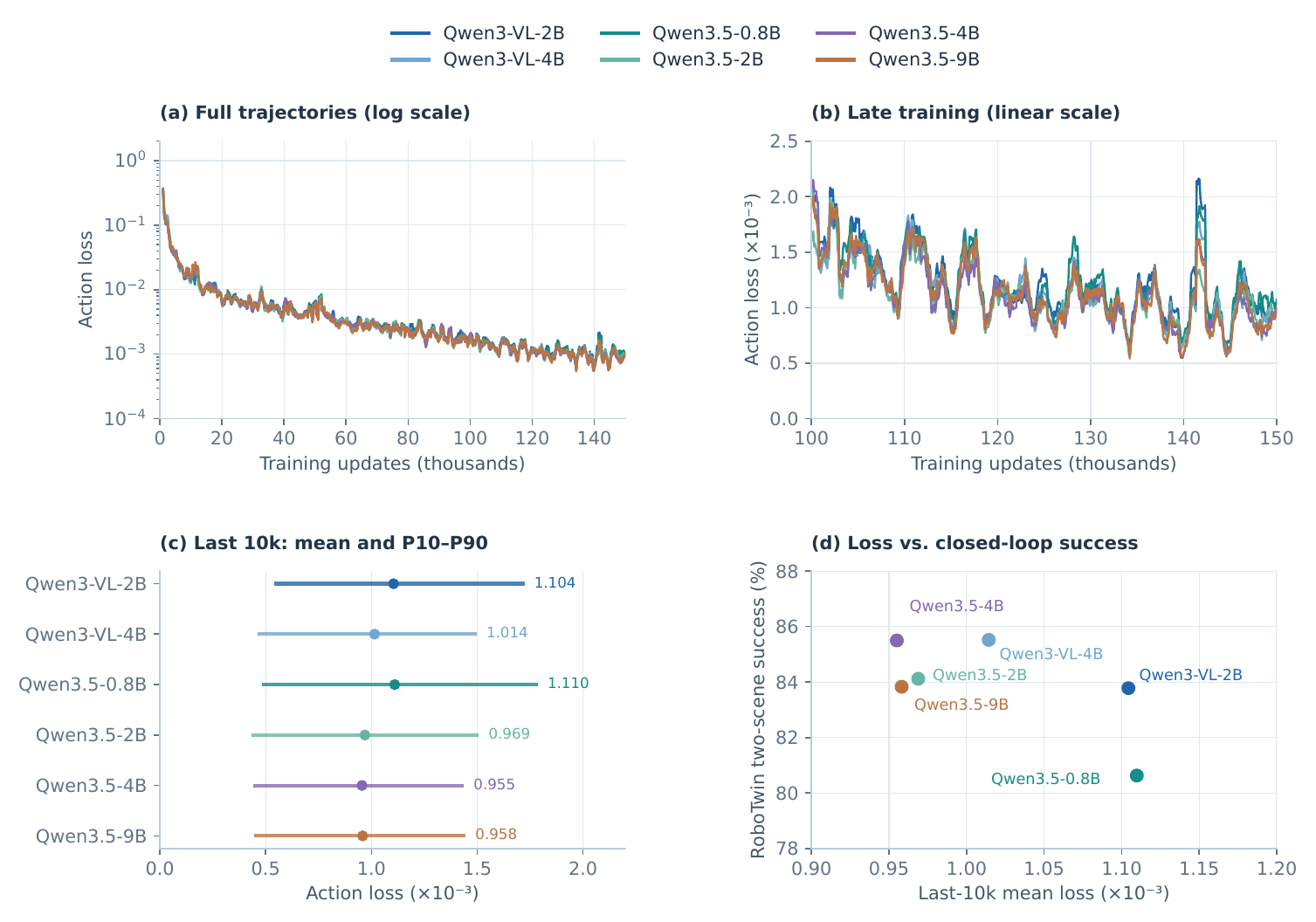}{1}{RoboTwin loss comparison and closed-loop outcomes. (a) All six trajectories use ten-observation moving averages on a common logarithmic scale over 0--150k updates. (b) The same smoothed series are enlarged over 100k--150k updates on a shared linear scale. (c) Dots show the mean of 100 raw observations with $140{,}000 < \mathrm{step} \leq 150{,}000$; horizontal lines span their empirical 10th--90th percentiles. These ranges describe within-run variation, not confidence intervals or variation across training seeds. (d) Tail means are paired with independently measured, equally weighted Clean/Randomized success rates for all six evaluated backbones. Lower training loss alone does not establish better task performance.}{fig:loss}

\textbf{Similar trajectory shapes can hide modest differences in final loss.} The full-run overlays share a rapid initial decline and continued late improvement. The last-10k means range from $0.955\times10^{-3}$ to $1.110\times10^{-3}$, so the smaller differences are difficult to see on a full-range logarithmic plot. The shared linear zoom exposes the late-stage levels and fluctuations without assigning each model a different vertical scale. Relative to the raw-observation mean over 100k--110k updates, last-10k means are about 24.7\%--32.6\% lower across the six runs, indicating continuing optimization rather than a fully flat tail.

\textbf{Within-run spread accompanies each mean.} The 10th--90th percentile ranges in panel (c) overlap substantially. They describe variation among logged training observations, which can reflect minibatch composition and stochastic training effects; they do not establish statistical equivalence or quantify variation across independently trained checkpoints. The raw observations remain available in the delivered CSV, and the derived means and percentiles are recorded in \texttt{robotwin\_loss\_statistics.csv}.

\textbf{Similar tail losses do not imply similar task success.} Qwen3.5-4B and Qwen3.5-9B have nearly identical last-10k mean losses ($0.955\times10^{-3}$ and $0.958\times10^{-3}$), yet their RoboTwin success rates are 85.50\% and 83.83\%. Conversely, Qwen3.5-0.8B and Qwen3-VL-2B have similar tail means near $1.11\times10^{-3}$ but different success rates, 80.63\% and 83.78\%. Panel (d) connects the optimization records to the task outcomes used for model selection; it does not fit a causal or statistically validated predictive relationship.

Intermediate-checkpoint rollouts complement loss monitoring. For Qwen3.5-4B, experimental records report Randomized success of 81.4\% at 110k and 85.8\% at 130k, using ten episodes per task. These small evaluations screen for capability changes during training. The final checkpoint reaches 85.12\% with 100 episodes per task. Differences between these evaluation sizes do not establish late-training degradation, particularly when individual ten-episode estimates are sensitive to sampling.

The 9B model finishes with action loss approximately $1.47\times10^{-3}$, while Qwen3.5-2B finishes near $1.44\times10^{-3}$. These final single observations differ from the last-10k means in panels (c,d). Qwen3.5-2B is now paired with its verified 84.12\% two-scene success rate in panel (d). Loss magnitudes from different datasets or methods are not pooled, since objectives, action representations, valid channels, and task distributions differ.

\subsubsection{Rollout Concurrency and Runtime}
\begin{table}[!htbp]\centering\small
\caption{Complete RoboTwin rollout details. Every configuration uses four evaluation GPUs. Clean and Randomized times are workflow wall times and may overlap.}\label{tab:rollout}
\setlength{\tabcolsep}{5pt}
\begin{tabular}{ccccc}\toprule
Backbone & Max. workers/GPU & Clean (h) & Randomized (h) & Episodes\\\midrule
Qwen3.5-0.8B & 6 & 4.15 & 5.52 & 10,000 \\
Qwen3-VL-2B & 6 & 4.26 & 4.88 & 10,000 \\
Qwen3-VL-4B & 4 & 6.38 & 6.77 & 10,000 \\
Qwen3.5-2B & 6 & 4.42 & 9.10 & 10,000 \\
Qwen3.5-4B & 4 & 8.68 & 11.31 & 10,000 \\
Qwen3.5-9B & 4 & 4.51 & 6.38 & 10,000 \\
\bottomrule\end{tabular}\end{table}

Evaluation separates policy serving from simulation and uses queues to limit concurrency. The evaluated 0.8B and 2B configurations allow up to six workers per GPU, while both 4B configurations and 9B allow four. Runtime therefore reflects backbone cost, scheduling, simulation, CPU contention, and I/O together. RoboTwin evaluation videos were disabled to reduce rendering and storage overhead, so a complete video archive is unavailable for these runs. LIBERO, LIBERO-plus, and RoboCasa retain video outputs used for aggregation or diagnosis.

Publishing these conditions helps readers distinguish workflow completion time from per-request speed. Future measurements will include policy-latency distributions, environment-step throughput, and execution rates under fixed hardware, inputs, and concurrency.

The 9B Clean and Randomized workflows take 4.51 and 6.38 hours, respectively. Their shorter elapsed times than the Qwen3.5-4B workflows do not establish faster 9B inference: the runs occurred under different software and shared-host conditions. A fixed-concurrency latency experiment is required to make that comparison.

Qwen3.5-2B Clean evaluation is recorded as 4:25:21. Its Randomized time of 9:05:43 sums the original queue and two repair queues (4:07:32, 2:47:13 and 2:10:58) following VGGT access failures affecting 19 tasks. Final scores use one complete result per task, rather than pooling failed attempts. This timing includes the repair burden and is not a matched estimate of successful-rollout speed.

\subsubsection{Task-Level Profiles and Uneven Scale Benefits}
\fig{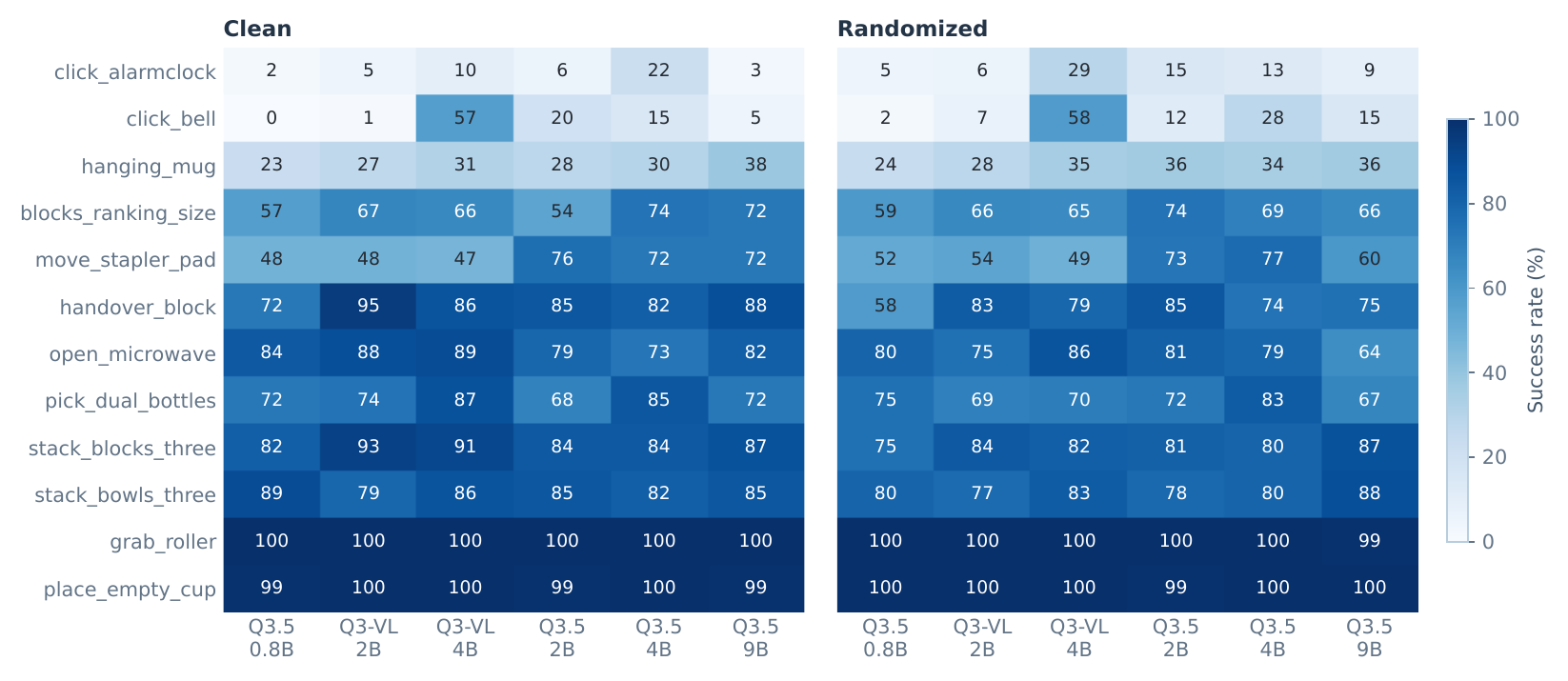}{1}{Success rates for selected tasks. The columns identify six evaluated configurations, including both 2B and both 4B backbones. Values are percentages. Precise contact, hanging, handover, and stacking tasks exhibit different responses to backbone scale. All 50 task counts appear in Appendix~\ref{app:tasks}.}{fig:task}

Scale benefits vary by task. Clean success on \texttt{click\_bell} increases from 0\%/1\% for the two small configurations to 57\% for Qwen3-VL-4B, while Qwen3.5-4B reaches 15\%. This motivates examining interactions between backbone family and task rather than relying solely on averages. Tasks such as \texttt{grab\_roller} are nearly saturated across scales, whereas \texttt{hanging\_mug} remains weak throughout.

Despite similar aggregate 4B scores, \texttt{move\_stapler\_pad}, \texttt{open\_microwave}, and \texttt{click\_bell} show substantial differences. These observations motivate diagnosis of data coverage, visual localization, and contact timing; they do not establish a particular module as the cause. Reported anomalies in action-dimension correlations are likewise diagnostic leads rather than proven controller-interface errors.

The 9B results further demonstrate uneven scale benefits. Compared with Qwen3.5-4B, \texttt{hanging\_mug} improves from 30/34 to 38/36 Clean/Randomized successes and \texttt{stack\_bowls\_three} improves from 82/80 to 85/88. In contrast, \texttt{click\_alarmclock} falls from 22/13 to 3/9 and \texttt{pick\_dual\_bottles} falls from 85/83 to 72/67. Larger capacity does not consistently remove the contact and manipulation bottlenecks in this recipe. Action-expert capacity, selected VLM feature layers, and data scale are plausible ablation axes, rather than established explanations.

\subsubsection{LIBERO and Zero-Shot Robustness}
\begin{table}[!htbp]\centering\small
\caption{ABot-M0 with Qwen3.5-4B on single-arm tasks and perturbations. LIBERO-plus uses the same 40k LIBERO checkpoint without additional training; denominators differ by suite.}\label{tab:libero}
\setlength{\tabcolsep}{6pt}
\begin{tabular}{ccccc}\toprule
& \multicolumn{2}{c}{LIBERO} & \multicolumn{2}{c}{LIBERO-plus}\\
Suite & Success / total & Rate (\%) & Success / total & Rate (\%)\\\midrule
Spatial & 487/500 & 97.40 & 2072/2402 & 86.26 \\
Object & 500/500 & 100.00 & 2100/2518 & 83.40 \\
Goal & 488/500 & 97.60 & 1953/2591 & 75.38 \\
Long & 460/500 & 92.00 & 1902/2519 & 75.51 \\
\midrule Overall micro-average & 1935/2000 & \textbf{96.75} & 8027/10030 & \textbf{80.03} \\
\bottomrule
\end{tabular}\end{table}

\fig{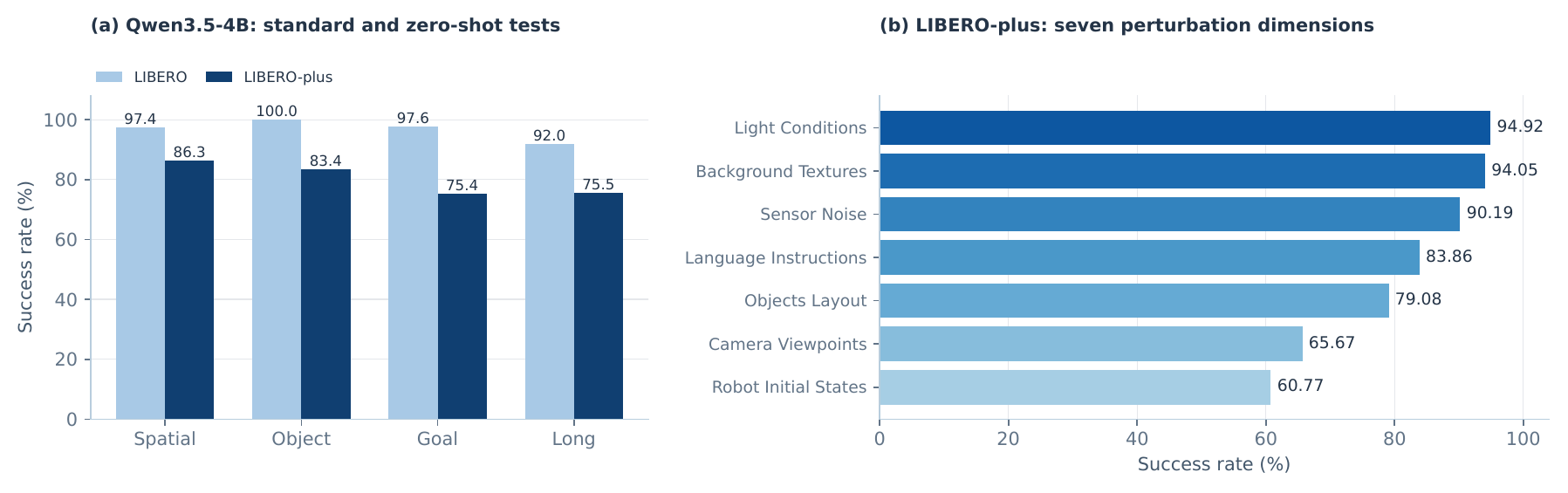}{1}{Performance by suite on LIBERO and LIBERO-plus, and by perturbation category on LIBERO-plus. LIBERO uses 50 episodes per task; LIBERO-plus uses one per variant. The two benchmarks are not pooled.}{fig:robustness}

Standard LIBERO success is 96.75\%, including 92.00\% on the Long suite. Without additional training, the micro-average across all 10,030 LIBERO-plus variants is 80.03\%. This exposes perturbation sensitivity beyond performance on the standard tasks.

\begin{table}[!htbp]\centering\small
\caption{LIBERO-plus perturbation categories, reaggregated from variant-level outcomes and task classifications.}\label{tab:perturb}
\begin{tabular}{ccc}\toprule
Category & Success / total & Rate (\%)\\\midrule
Background Textures & 1012/1076 & 94.05 \\
Camera Viewpoints & 1050/1599 & 65.67 \\
Language Instructions & 1289/1537 & 83.86 \\
Light Conditions & 1084/1142 & 94.92 \\
Objects Layout & 1206/1525 & 79.08 \\
Robot Initial States & 942/1550 & 60.77 \\
Sensor Noise & 1444/1601 & 90.19 \\
\bottomrule\end{tabular}\end{table}

Robot initial states and camera viewpoints yield 60.77\% and 65.67\%, below lighting (94.92\%) and background textures (94.05\%). The gaps motivate improvements to geometric reasoning and state conditioning. However, categories differ in perturbation difficulty and composition; the comparison does not establish causality or demonstrate understanding of all language variations.

\textbf{Deduplication and result transparency.} Original outputs contain duplicates for 101 tasks, including three tasks with conflicting success labels. We use $(\mathrm{suite},\mathrm{task\ id})$ as the unique key and retain the output with the latest modification time as the rerun record, obtaining 8,027 successes. Assigning all three conflicting tasks either failure or success bounds the overall score at 80.01--80.04\%. The selected records and all alternatives are preserved in the delivered audit data. The rerun-selection rule should be checked against run logs before formal publication. With only one episode per variant, this evaluation does not measure repeat-execution stability for individual variants.

\subsubsection{RoboCasa: High-Dimensional Control and Failure Diagnosis}
RoboCasa GR1 Tabletop uses 29-dimensional absolute joint targets: seven dimensions per arm, six per hand, and three for the waist. Evaluation covers 24 tasks with 50 episodes each, totaling 1,200 rollouts. Each episode allows up to 720 environment steps, and each prediction executes 12 action steps. Records report approximately 20 hours 44 minutes for 50k training updates on four H200 GPUs, averaging approximately 1.405 seconds per update. This timing is an approximate recorded statistic, unlike the exact RoboTwin wall-time values.

The policy succeeds in 574/1200 episodes (47.83\%). Task scores range from 16\% to 76\%, with lower success on tiered shelves and some placement-and-closure tasks. Transferring an object from a cutting board to a pan reaches 76\%.

\fig{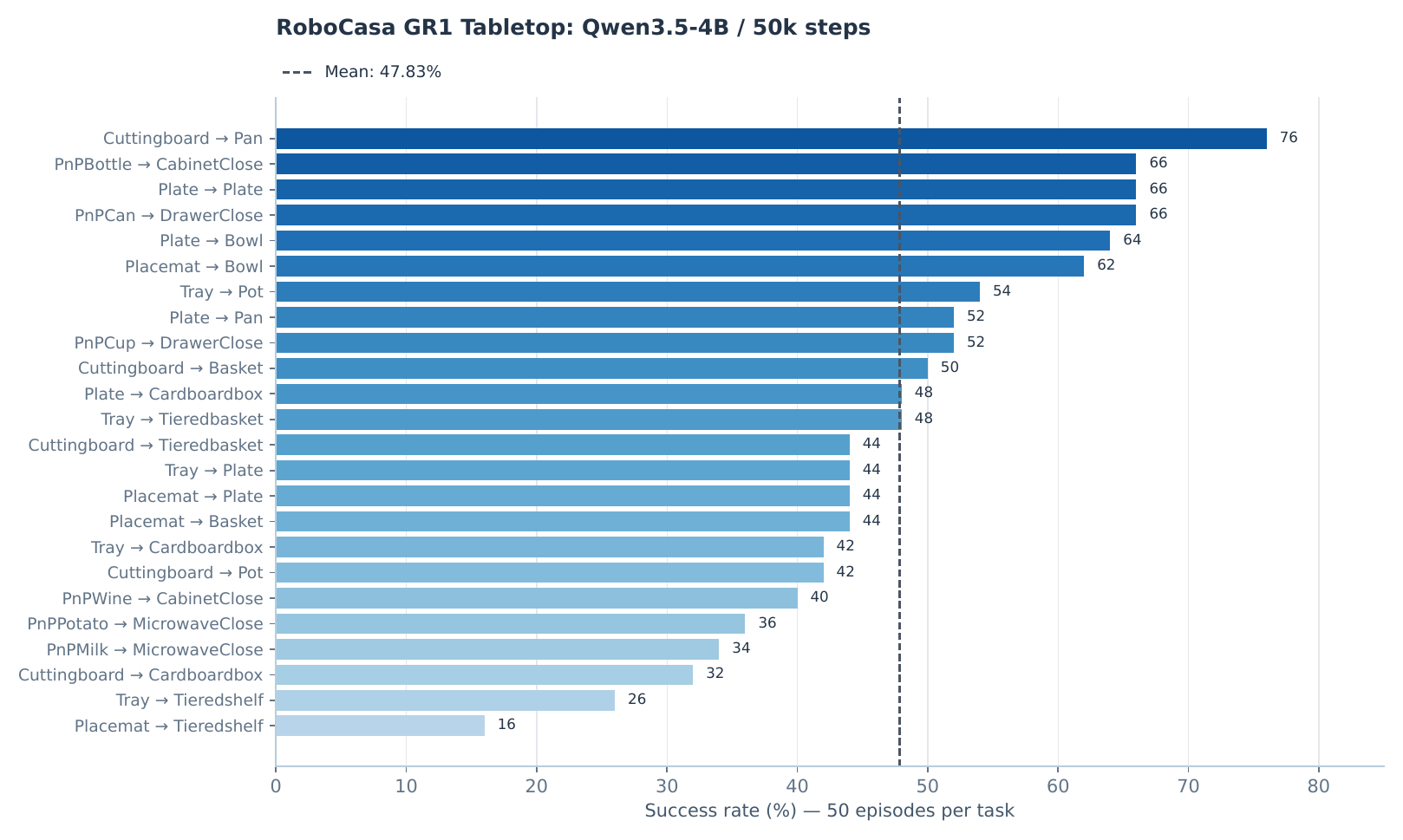}{0.98}{All 24 RoboCasa GR1 Tabletop tasks. The dashed line marks mean success of 47.83\%. Equal counts of 50 episodes per task make macro- and micro-averages identical. Figure labels are abbreviated; full task names are retained in the data tables.}{fig:robocasa}

\textbf{Process evidence supports targeted diagnosis.} Existing failure videos and probe records show rollouts that fail to approach an object before grasping, sweep it away, or continue until timeout after an unsuccessful grasp. In one tracked sequence, the end effector remains approximately 0.2--0.3 m from the target and the object barely moves, while commanded and measured joint angles differ by approximately 0.01--0.06 rad. For these observed trajectories, the evidence motivates checking the generated policy actions before attributing failure to controller tracking.

The experimenters checked action ordering, normalization statistics, absolute-joint semantics, and state encoding. An early loss plateau and outputs close to current state are consistent with a shortcut that exploits temporal continuity. \textbf{The inertial-shortcut and low-learning-rate explanations remain hypotheses}: probes do not cover every failure, and no controlled learning-rate ablation establishes causality. Follow-up experiments should vary learning rates and state conditions, measure dependence on visual input, and add periodic closed-loop monitoring.

\subsubsection{CALVIN: Continuous Language-Conditioned Task Chains}\label{sec:calvin}
\textbf{Training and interfaces.} ABot-M0 with Qwen3.5-4B is trained independently on demonstrations from CALVIN environments A/B/C/D and evaluated in environment D. The converted LeRobot v2.1 dataset contains 17,870 episodes and 1,071,743 frames, with a $200\times200$ static camera and an $84\times84$ wrist camera. Images are resized for the policy interface. This ABCD $\rightarrow$ D setting includes D in training; it does not test transfer to a held-out environment D.

Training uses four H200 GPUs, global batch 64, 34,000 updates, seed 42, frozen VGGT, and a DiT-B action expert. Base/VLM/action learning rates are $2.5\times10^{-5}$, $10^{-5}$, and $10^{-4}$, with 5,000 warmup updates and a cosine schedule. The action horizon is eight. CALVIN's seven action channels (relative end-effector translation, rotation, and gripper) occupy the right-arm half of the 14-dimensional internal representation; the configured state layout is 16-dimensional after padding. The saved training script requests \texttt{include\_state=false}, and the inspected rollout client sends camera images and language without an explicit robot-state field. Configured state dimensions therefore do not establish active proprioceptive conditioning.

The final run completes in 14:39:51 (14.66 hours; 58.7 GPU-hours), with approximately 41.2 nominal samples/s and a 13.27 GB final checkpoint. Its last logged action loss is 0.02949. The console file also contains preliminary runs; only the final uninterrupted 340 observations are used for its loss curve in Appendix~\ref{app:calvin}. Preliminary observations and the selection rule are preserved in the audit.

\textbf{Evaluation protocol.} Evaluation uses the stored list of 1,000 five-task sequences and CALVIN's task oracle. The environment resets to the prescribed initial condition at the start of a sequence and carries physical state between successful subtasks; the first failure terminates the chain. Each subtask has a 360-step limit. A single simulation client communicates with a separate policy server. In the inspected client, the executed chunk length is read from the checkpoint configuration and equals eight; the wrapper's \texttt{replan\_steps=5} field does not control this interval. Policy buffers reset between subtasks. Evaluation wall time is recorded as approximately 10 hours 16 minutes, rather than a fixed-hardware latency measurement.

\begin{table}[!htbp]\centering\small
\caption{CALVIN ABCD $\rightarrow$ D: successful prefixes for the final 34k Qwen3.5-4B checkpoint. Every row uses the same 1,000 sequences.}\label{tab:calvin_chain}
\begin{tabular}{ccc}\toprule
Consecutive tasks completed & Sequence count & Rate (\%)\\\midrule
$\geq 1$ & 889/1000 & 88.9 \\
$\geq 2$ & 798/1000 & 79.8 \\
$\geq 3$ & 721/1000 & 72.1 \\
$\geq 4$ & 646/1000 & 64.6 \\
$\geq 5$ & 571/1000 & 57.1 \\
\midrule
\multicolumn{2}{c}{Average successful sequence length (maximum 5)} & \textbf{3.625}\\
\bottomrule\end{tabular}\end{table}

The model completes at least one task in 88.9\% of sequences and all five in 57.1\%. The five prefix counts sum to 3,625 successes, giving $3,625/1,000=3.625$. The task table independently sums to the same 3,625 successes across 4,054 reached subtask attempts. The latter denominator includes the 429 failures that terminate incomplete chains, rather than all 5,000 scheduled task positions. Sequence-level composition and conditional task success must therefore remain distinct.

\fig{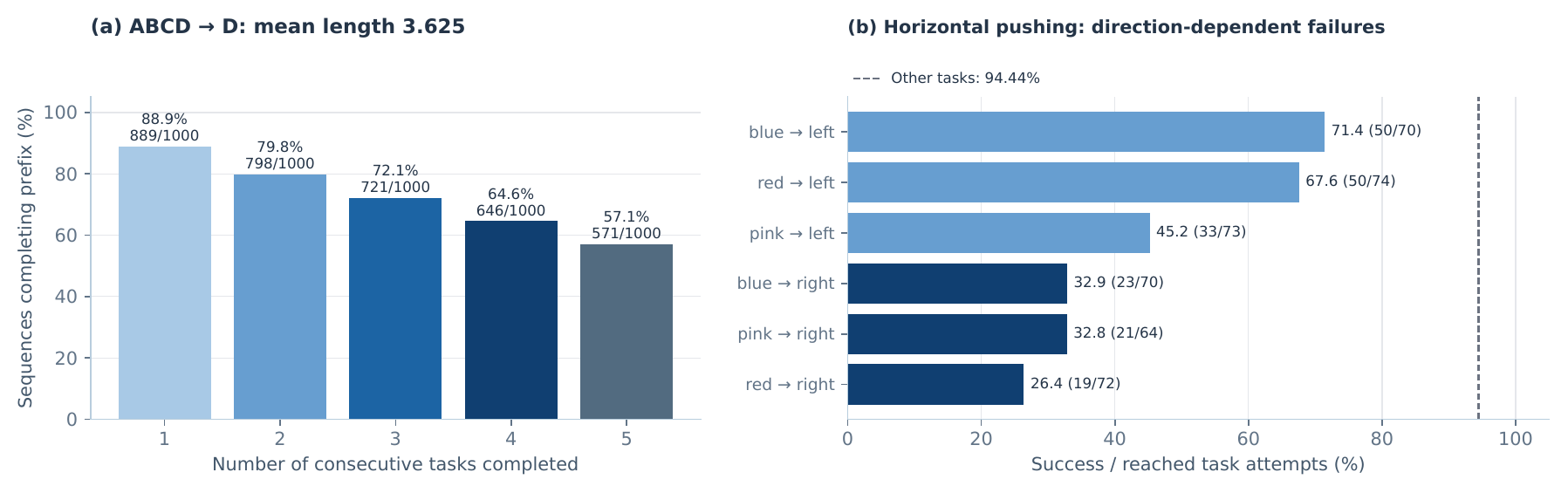}{1}{CALVIN chain completion and horizontal pushing. The left panel uses all 1,000 sequences as its denominator. The right panel uses actual reached attempts for each push task; the dashed reference aggregates the other 28 task types. Task frequencies and positions differ, so the reference is diagnostic rather than a matched causal comparison.}{fig:calvin}

\textbf{Horizontal pushing is the main observed bottleneck.} Six left/right block-pushing tasks account for 196/423 successes (46.34\%), while the other 28 task types account for 3,429/3,631 (94.44\%). Rightward pushes succeed in 63/206 attempts (30.58\%), compared with 133/217 (61.29\%) for leftward pushes. In total, 26 of the 34 task types reach at least 85\% conditional success. \texttt{push\_into\_drawer} is retained among the other tasks because the horizontal block-pushing group is explicitly defined by the six color/direction tasks.

\begin{table}[!htbp]\centering\small
\caption{CALVIN failure groups. Push left and Push right partition Horizontal push; Other tasks form its complement. Rates are conditional on reaching a task.}\label{tab:calvin_families}
\begin{tabular}{ccc}\toprule
Task group & Success / attempts & Rate (\%)\\\midrule
Horizontal push & 196/423 & 46.34 \\
Push left & 133/217 & 61.29 \\
Push right & 63/206 & 30.58 \\
Other tasks & 3429/3631 & 94.44 \\
\bottomrule\end{tabular}\end{table}

The direction difference motivates checking demonstration counts, signed action distributions, initial object positions, and normalization statistics, followed by failure replay and action-horizon ablations. Contact loss during an eight-step open-loop chunk is another candidate explanation. No controlled experiment yet isolates these effects, and the aggregate rates do not justify a numerical prediction of improvement after fixing pushing.

\textbf{Result selection and evidence quality.} The source JSON contains two occurrences of the same top-level key: an earlier ten-sequence smoke test and the final complete evaluation. We select the final occurrence and preserve both records in \texttt{calvin\_audit.json}; the smoke test is not pooled. The narrative source table omits \texttt{close\_drawer}, repeats \texttt{lift\_pink\_block\_drawer}, and overstates the number of tasks above 85\%. The delivered table follows all 34 JSON task keys. Chain and task totals are internally consistent, but a per-sequence outcome file and a complete failure-video archive are unavailable in the inspected artifacts. The stored sequence list and evaluation code are hashed for provenance; external score comparisons should additionally match evaluator versions and execution settings.

\subsubsection{Interpretation and Evidence Limits}
The results support discussions of scale trade-offs under the specified ABot-M0 recipe and simulation settings. The following limits govern their interpretation:
\begin{itemize}
\item All six RoboTwin configurations have complete two-scene scores, including the newly audited Qwen3.5-2B result.
\item Each completed configuration uses one training seed. One hundred episodes per task characterize that checkpoint, not variation across training runs.
\item Model scales refer to nominal VLM sizes. Full policies also contain spatial modules, action experts, and projections whose sizes may change with hidden width.
\item Update count, batch, and data are controlled, while backbone pre-train histories, dependency versions, and kernel paths remain systematic differences.
\item Memory lacks a uniform measurement protocol, and rollout time depends on concurrency and simulation load. Neither establishes minimum deployment memory or a strict latency advantage.
\item Action spaces, training steps, and learning rates differ across datasets. The cross-dataset scores do not form a single scaling law. This report includes no real-robot results.
\item CALVIN measures successful prefixes of continuous task chains. Its ABCD training includes D; conditional task rates and sequence length do not constitute a held-out-environment transfer score. Available outcomes are aggregate records rather than independently audited per-sequence traces.
\end{itemize}
These evidence gaps should accompany performance and weights so readers can identify what supports each conclusion and what measurements remain necessary.

\input{sections/reproductions.tex}

\input{sections/assessment.tex}
\input{sections/px_harness.tex}

\section{Conclusion}
RoboSeed is a continuing effort to train, evaluate, and openly release robot models for different computing budgets, together with verifiable evidence for model selection. This report connects that community goal to local training, checkpoint evaluation, resource measurements and task-level diagnosis.

We propose RoboSeed as a multidimensional evaluation framework for community researchers selecting and studying robot action models. Its five dimensions connect task capability with downstream training affordability, deployment affordability, locally verified research value and reproduction evidence. Fixed anchors, explicit criteria, alternative preference settings and transparent missing-data rules turn heterogeneous experimental records into inspectable decision profiles. The framework helps researchers choose feasible configurations, assess the evidence behind mechanism claims and prioritize measurements that could change a decision.

The VLA and WAM studies instantiate this framework across four policy families and 12 checkpoint units. Within ABot-M0, the tested smaller backbones retain substantial success with lower training cost, while the completed 9B run shows that higher cost does not ensure better performance under a fixed-update recipe. Task-level robustness and continuous-chain evaluation further reveal limitations hidden by an overall success percentage. These empirical studies supply the evidence from which the framework derives budget-specific assessments.

The extension broadens the evidence to four policy families. OpenWAM local adaptation reaches 91.35\% across both RoboTwin scenes using a recorded 262.068 downstream H200 GPU-hours. LingBot's 89.72\% official-weight Clean result and 57.78\% locally adapted result reveal a large recipe gap requiring controlled investigation. LiLa-WAM official/local stage-2 weights reach 89.60\%/88.80\% Clean success using five episodes per task and visual task conditions. The 0.80-point observed gap does not establish statistical equivalence. PX demonstrates approximately twofold evaluation throughput improvement for similar workloads at equal concurrency, showing the independent value of deployment engineering.

The five-dimensional assessment makes each judgment explicit: raw capability, scoped training cost, deployment measurements, locally established research value, and evidence completeness. Dashes identify missing measurements and incomplete composite scores, while weighted coverage describes available inputs. The evidence supports budget-specific ABot choices and concrete cross-method follow-ups, while absent serving traces, single-seed training and protocol differences prevent a defensible universal winner. Reproducible formulas, source snapshots and machine-readable scores make those limits reviewable and future updates straightforward.

\section{Future Work}
\textbf{Close decision-critical gaps.} Measure warmed batch-one policy p50/p95 latency and peak VRAM on a common H200 protocol, recording native action horizons and all required modules. Complete LingBot/LiLa Randomized evaluation and expand matched LiLa sampling beyond five episodes per task. Audit LiLa's stage-2 learning-rate floor before testing a corrected schedule. These measurements complete missing scoring inputs and enable aggregate assessment.

\textbf{Test mechanisms under aligned budgets.} Compare LingBot exposure and batch schedules without promising an extrapolated success rate. For OpenWAM, isolate predictive supervision and world-to-action interaction while fixing the adaptation budget. Harmonize accepted seeds, instruction sampling, action execution and numerical precision before cross-method attribution, and repeat component ablations across at least three training seeds. Reassess the score policy only with a versioned change to its declared preferences.

\textbf{Expand backbones and scales.} Use the now-complete Qwen3.5-2B evaluation to refine budget choices, then add smaller, larger, and additional VLM families. Use the observed 9B result to test changes in action-expert capacity, feature selection, data coverage, and training duration through controlled ablations. Complement fixed-update comparisons with fixed-GPU-hour studies and multiple training seeds to distinguish budget effects from run variation.

\textbf{Expand datasets and methods.} Evaluate more tasks, embodiments, and data scales, and extend replaceable interfaces to additional VLA methods. Preserve checks of dataset semantics, action conventions, and evaluator versions, while reporting adaptation cost and task performance separately.

\textbf{Diagnose CALVIN composition failures.} Audit the six horizontal-pushing task distributions and signed action statistics, retain per-sequence outcomes and representative failure videos, and compare shorter and longer action chunks under the same sequence list. Evaluate intermediate checkpoints and multiple training seeds before estimating gains. Separate ABCD $\rightarrow$ D from held-out-environment settings such as ABC $\rightarrow$ D.

\textbf{Rapidly evaluate community techniques.} Establish a workflow from compatibility checks and short training runs through screening rollouts and complete evaluation. Verify active acceleration kernels when assessing new architectures.

\textbf{Strengthen process measurements.} Standardize memory sampling across allocated, reserved, and device monitoring readings, with explicit units, peak windows, and device capacity. Publish training-stage runtimes, losses, and anomalies. Record policy p50/p95 latency, simulation throughput, success counts, exclusions, and retries under fixed concurrency, alongside representative success and failure videos.

\textbf{Connect model releases to continuing report updates.} We plan to publish locally trained checkpoints and associated experimental results on Hugging Face, linked from the project repository at \url{https://github.com/PrimitiveX/RoboSeed}. Each model release should include a model card, effective configuration, normalization statistics, dataset and implementation versions, checksums, and the corresponding evaluation records. Released-weight evaluation and local training outcomes will remain distinct, and unevaluated checkpoints will retain their pending status. Verified download entries and immutable revisions will accompany actual publication. Subsequent report versions will document new models, completed measurements and any resulting changes to selection guidance. Real-robot experiments will test whether the observed simulation trade-offs carry over to deployment.

\appendix
\section{Experimental Details and Evidence Completeness}\label{app:details}
\subsection{Dependencies, Interfaces, and Evaluation Settings}
Recorded RoboTwin training dependencies are Python 3.12.3, PyTorch 2.8.0+cu128, accelerate 1.5.2, DeepSpeed ZeRO-2, and bf16 without CPU offload. Qwen3-VL uses transformers 4.57.0, and Qwen3.5 uses 5.15.1. These are run-specific records; reproduction should pin compatible combinations. Effective configuration snapshots saved during training take precedence over repository defaults modified afterward.

\begin{table}[!htbp]\centering\small
\caption{Key interfaces and evaluation settings. LIBERO-plus entries describe this script snapshot rather than a universal protocol for other implementations.}
\begin{tabularx}{\linewidth}{C{3.0cm}YYY}\toprule
Setting & RoboTwin & LIBERO / LIBERO-plus & RoboCasa GR1\\\midrule
Training updates & 150k & 40k / no extra training & 50k\\\addlinespace[3pt]
Global batch & 64 & 64 & 64\\\addlinespace[3pt]
State / action dimensions & 14 / 14 & 7 / 14 in config; state input to be verified & 58 / 29\\\addlinespace[3pt]
Training action horizon & 50 & 10 & 16\\\addlinespace[3pt]
Integration steps & 4 & 4 & 4\\\addlinespace[3pt]
Training seed & 42 & 42 & 42\\\addlinespace[3pt]
Evaluation seed & Launch argument 0 & Script default 7 & Not recorded in summary\\\addlinespace[3pt]
Episodes per task & 100 per scene & 50 / 1 per variant & 50\\\addlinespace[3pt]
Environment step limit & Task dependent & 220 / 280 / 300 / 520 by suite; 10 settling steps & 720\\\addlinespace[3pt]
Rollout videos & Disabled & Available & Available\\\bottomrule
\end{tabularx}\end{table}

The LIBERO-plus script inherits the four suite-specific step limits and settling period. Reproduction and external comparisons should check these settings, executed action chunks, random states, and environment versions. Complete task coverage alone does not imply equivalence to every public evaluation protocol.

\subsection{Available Evidence and Remaining Measurements}
\begin{table}[!htbp]\centering\small
\caption{ABot evidence completeness after the October 28 refresh. Additional method results are audited separately in Appendix~\ref{app:extension}.}
\begin{tabularx}{\linewidth}{C{3.05cm}YY}\toprule
Evidence & Current status & Follow-up\\\midrule
RoboTwin success & Original results checked against final log counters; 50 tasks per scene & Multiple training seeds and external protocol checks\\\addlinespace[3pt]
Configurations and final weights & Six evaluated RoboTwin configurations and dataset-specific final weights; file sizes measured & Versioned release configurations and weight checksums\\\addlinespace[3pt]
Loss convergence & Observations extracted from console logs & Standard exports of raw metrics\\\addlinespace[3pt]
Training memory & Approximate recorded mean and maximum readings & Raw traces, units, and capacity checks\\\addlinespace[3pt]
Rollout time and concurrency & Workflow times and concurrency limits recorded & Policy latency, task time, and worker activity\\\addlinespace[3pt]
LIBERO / RoboCasa counts & Reconstructed video verdicts and task denominators & Independent episode tables and exception logs\\\addlinespace[3pt]
LIBERO-plus coverage and duplicates & All variants covered; duplicates and conflicts preserved & Confirm rerun replacement using run IDs and logs\\\addlinespace[3pt]
CALVIN chains and task counts & Final aggregate record selected; 34 task types and chain totals checked & Per-sequence traces, evaluator commit, and failure videos\\\addlinespace[3pt]
Public repository and weight entries & Project repository provided; release materials under expansion & Weight download links and dependency documentation\\\bottomrule
\end{tabularx}\end{table}

\subsection{Reproducing the Data and Figures}
The delivered \texttt{data/} directory contains six completed RoboTwin configurations and task tables, six training-loss series, task outcomes for LIBERO, LIBERO-plus, and RoboCasa, and LIBERO-plus category statistics and duplicate audits. CALVIN adds prefix counts, all 34 conditional task counts, failure-group statistics, final-run loss observations, the original aggregate JSON, and a duplicate-key selection audit. The refreshed \texttt{source\_manifest.json} records source paths and evidence hashes. Separate extension records contain source manifests, task counts, checkpoint metadata and score profiles; Appendix~\ref{app:extension} gives their audit and regeneration procedure. Original directory aliases appear only in provenance materials; method names in the report remain consistent. ABot video outcomes are reconstructed from environment-success labels in filenames, rather than independently relabeled through frame-by-frame inspection of every rollout.

The read-only \texttt{scripts/collect\_data.py} extracts evidence and checks coverage and denominators. The scripts \texttt{make\_tables.py} and \texttt{draw\_figures.py} generate manuscript tables and figures from shared data. Figures are exported as PDF, SVG, and PNG; framework diagrams remain editable vectors. All new illustrations use blue, teal, white, and muted gray. Original experiment files and weights are preserved.

\section{Complete RoboTwin Task Counts}\label{app:tasks}
Every numeric column reports successes out of 100 episodes, numerically equal to success rate in percent. C and R denote Clean and Randomized. Q3 denotes Qwen3-VL and Q3.5 denotes Qwen3.5. All six backbone pairs have complete evaluations, including Qwen3.5-2B.
\begingroup\scriptsize
\renewcommand{\arraystretch}{1.04}
\setlength{\tabcolsep}{2pt}
\begin{longtable}{lrrrrrrrrrrrr}
\caption{Success counts for all 50 RoboTwin tasks.}\label{tab:alltasks}\\
\toprule
& \multicolumn{2}{c}{Q3.5-0.8B} & \multicolumn{2}{c}{Q3-2B} & \multicolumn{2}{c}{Q3-4B} & \multicolumn{2}{c}{Q3.5-2B} & \multicolumn{2}{c}{Q3.5-4B} & \multicolumn{2}{c}{Q3.5-9B}\\
Task & C & R & C & R & C & R & C & R & C & R & C & R\\\midrule\endfirsthead
\toprule Task & C & R & C & R & C & R & C & R & C & R & C & R\\\midrule\endhead
\bottomrule\endfoot
adjust\_bottle & 100 & 100 & 100 & 100 & 100 & 100 & 100 & 99 & 100 & 99 & 100 & 100 \\
beat\_block\_hammer & 83 & 82 & 94 & 88 & 94 & 92 & 86 & 83 & 89 & 83 & 96 & 87 \\
blocks\_ranking\_rgb & 91 & 93 & 93 & 93 & 92 & 94 & 90 & 84 & 98 & 91 & 94 & 90 \\
blocks\_ranking\_size & 57 & 59 & 67 & 66 & 66 & 65 & 54 & 74 & 74 & 69 & 72 & 66 \\
click\_alarmclock & 2 & 5 & 5 & 6 & 10 & 29 & 6 & 15 & 22 & 13 & 3 & 9 \\
click\_bell & 0 & 2 & 1 & 7 & 57 & 58 & 20 & 12 & 15 & 28 & 5 & 15 \\
dump\_bin\_bigbin & 97 & 93 & 97 & 97 & 96 & 97 & 97 & 96 & 97 & 99 & 96 & 99 \\
grab\_roller & 100 & 100 & 100 & 100 & 100 & 100 & 100 & 100 & 100 & 100 & 100 & 99 \\
handover\_block & 72 & 58 & 95 & 83 & 86 & 79 & 85 & 85 & 82 & 74 & 88 & 75 \\
handover\_mic & 100 & 99 & 100 & 100 & 98 & 98 & 100 & 100 & 100 & 99 & 98 & 99 \\
hanging\_mug & 23 & 24 & 27 & 28 & 31 & 35 & 28 & 36 & 30 & 34 & 38 & 36 \\
lift\_pot & 99 & 100 & 100 & 100 & 100 & 100 & 100 & 99 & 100 & 98 & 100 & 100 \\
move\_can\_pot & 83 & 71 & 87 & 83 & 84 & 74 & 82 & 84 & 82 & 86 & 73 & 75 \\
move\_pillbottle\_pad & 93 & 86 & 93 & 92 & 96 & 95 & 94 & 92 & 93 & 90 & 96 & 94 \\
move\_playingcard\_away & 97 & 95 & 97 & 98 & 100 & 98 & 99 & 98 & 99 & 100 & 97 & 99 \\
move\_stapler\_pad & 48 & 52 & 48 & 54 & 47 & 49 & 76 & 73 & 72 & 77 & 72 & 60 \\
open\_laptop & 95 & 98 & 98 & 99 & 97 & 99 & 95 & 100 & 97 & 100 & 97 & 98 \\
open\_microwave & 84 & 80 & 88 & 75 & 89 & 86 & 79 & 81 & 73 & 79 & 82 & 64 \\
pick\_diverse\_bottles & 41 & 59 & 69 & 67 & 71 & 59 & 65 & 69 & 63 & 61 & 64 & 57 \\
pick\_dual\_bottles & 72 & 75 & 74 & 69 & 87 & 70 & 68 & 72 & 85 & 83 & 72 & 67 \\
place\_a2b\_left & 81 & 90 & 88 & 87 & 93 & 96 & 92 & 93 & 95 & 97 & 93 & 94 \\
place\_a2b\_right & 78 & 83 & 86 & 83 & 91 & 92 & 90 & 91 & 91 & 88 & 95 & 94 \\
place\_bread\_basket & 85 & 81 & 85 & 81 & 87 & 77 & 86 & 85 & 92 & 86 & 91 & 86 \\
place\_bread\_skillet & 89 & 81 & 84 & 83 & 82 & 89 & 89 & 77 & 94 & 86 & 91 & 88 \\
place\_burger\_fries & 99 & 98 & 96 & 98 & 95 & 98 & 100 & 100 & 99 & 95 & 99 & 98 \\
place\_can\_basket & 70 & 63 & 75 & 76 & 79 & 75 & 73 & 77 & 81 & 72 & 70 & 72 \\
place\_cans\_plasticbox & 100 & 92 & 100 & 96 & 100 & 99 & 100 & 97 & 98 & 96 & 99 & 98 \\
place\_container\_plate & 100 & 99 & 99 & 100 & 98 & 99 & 100 & 100 & 100 & 100 & 99 & 100 \\
place\_dual\_shoes & 85 & 82 & 80 & 79 & 82 & 84 & 91 & 83 & 84 & 85 & 75 & 81 \\
place\_empty\_cup & 99 & 100 & 100 & 100 & 100 & 100 & 99 & 99 & 100 & 100 & 99 & 100 \\
place\_fan & 94 & 91 & 96 & 98 & 99 & 97 & 93 & 92 & 94 & 97 & 98 & 94 \\
place\_mouse\_pad & 68 & 68 & 75 & 78 & 73 & 70 & 77 & 77 & 82 & 73 & 76 & 73 \\
place\_object\_basket & 88 & 88 & 92 & 89 & 88 & 86 & 97 & 86 & 89 & 88 & 90 & 86 \\
place\_object\_scale & 88 & 80 & 93 & 93 & 95 & 90 & 96 & 86 & 94 & 93 & 92 & 94 \\
place\_object\_stand & 96 & 88 & 90 & 91 & 94 & 86 & 94 & 91 & 97 & 92 & 97 & 92 \\
place\_phone\_stand & 77 & 93 & 87 & 98 & 94 & 97 & 91 & 95 & 91 & 96 & 91 & 99 \\
place\_shoe & 94 & 96 & 96 & 97 & 95 & 100 & 91 & 100 & 94 & 97 & 93 & 99 \\
press\_stapler & 95 & 94 & 98 & 97 & 93 & 94 & 95 & 93 & 96 & 94 & 87 & 88 \\
put\_bottles\_dustbin & 67 & 79 & 79 & 75 & 75 & 80 & 82 & 74 & 79 & 89 & 74 & 81 \\
put\_object\_cabinet & 91 & 91 & 96 & 96 & 87 & 90 & 92 & 86 & 91 & 92 & 90 & 95 \\
rotate\_qrcode & 90 & 90 & 90 & 93 & 91 & 94 & 84 & 88 & 91 & 94 & 84 & 91 \\
scan\_object & 74 & 83 & 78 & 88 & 89 & 87 & 83 & 81 & 81 & 81 & 79 & 77 \\
shake\_bottle & 100 & 100 & 100 & 100 & 100 & 100 & 100 & 98 & 98 & 100 & 100 & 100 \\
shake\_bottle\_horizontally & 100 & 99 & 100 & 100 & 100 & 99 & 99 & 99 & 100 & 99 & 100 & 100 \\
stack\_blocks\_three & 82 & 75 & 93 & 84 & 91 & 82 & 84 & 81 & 84 & 80 & 87 & 87 \\
stack\_blocks\_two & 100 & 97 & 98 & 99 & 96 & 99 & 95 & 96 & 98 & 93 & 100 & 98 \\
stack\_bowls\_three & 89 & 80 & 79 & 77 & 86 & 83 & 85 & 78 & 82 & 80 & 85 & 88 \\
stack\_bowls\_two & 98 & 97 & 96 & 98 & 96 & 98 & 98 & 96 & 99 & 98 & 99 & 97 \\
stamp\_seal & 64 & 65 & 68 & 86 & 75 & 86 & 68 & 78 & 78 & 83 & 74 & 78 \\
turn\_switch & 60 & 71 & 53 & 70 & 60 & 63 & 67 & 68 & 71 & 69 & 49 & 67 \\
\end{longtable}\endgroup

\section{CALVIN Configuration, Training, and Complete Task Counts}\label{app:calvin}
\begin{table}[!htbp]\centering\small
\caption{CALVIN reproduction details from the saved configuration and inspected evaluation interfaces.}\label{tab:calvin_setup}
\begin{tabularx}{\linewidth}{C{4.05cm}Y}\toprule
Setting & Recorded value\\\midrule
Training / evaluation environments & ABCD / D\\\addlinespace[3pt]
Demonstrations / frames & 17,870 / 1,071,743; converted LeRobot v2.1\\\addlinespace[3pt]
Updates / batch / seed & 34,000 / 64 / 42\\\addlinespace[3pt]
Frozen encoder & VGGT spatial module\\\addlinespace[3pt]
Action expert / integration steps & DiT-B, 16 layers / configured 4\\\addlinespace[3pt]
Training / executed action horizon & 8 / 8 in the inspected shared client\\\addlinespace[3pt]
State / action layout & Configured 16 / 14; active CALVIN action channels: 7\\\addlinespace[3pt]
Policy observations & Static and wrist cameras plus language; no explicit robot-state field\\\addlinespace[3pt]
Normalization & Saved statistics, min--max action unnormalization, right-arm channel selection\\\addlinespace[3pt]
Subtask step limit / stop rule & 360 / terminate chain at first failure\\\addlinespace[3pt]
Evaluation clients / videos & One simulation client / debug video output disabled by default\\\addlinespace[3pt]
Training / recorded evaluation time & 14:39:51 / approximately 10:16:00\\\bottomrule
\end{tabularx}\end{table}

The saved configuration requests a minimum learning rate of $10^{-6}$, whereas the last logged scalar learning rate is $4\times10^{-7}$. Scheduler and parameter-group conventions must be reconciled before claiming an exact effective rate. The source notes describe acceleration dependencies for this later training environment; a matched kernel ablation has not been performed. These details accompany the result without implying a measured speedup relative to earlier 4B runs.

\fig{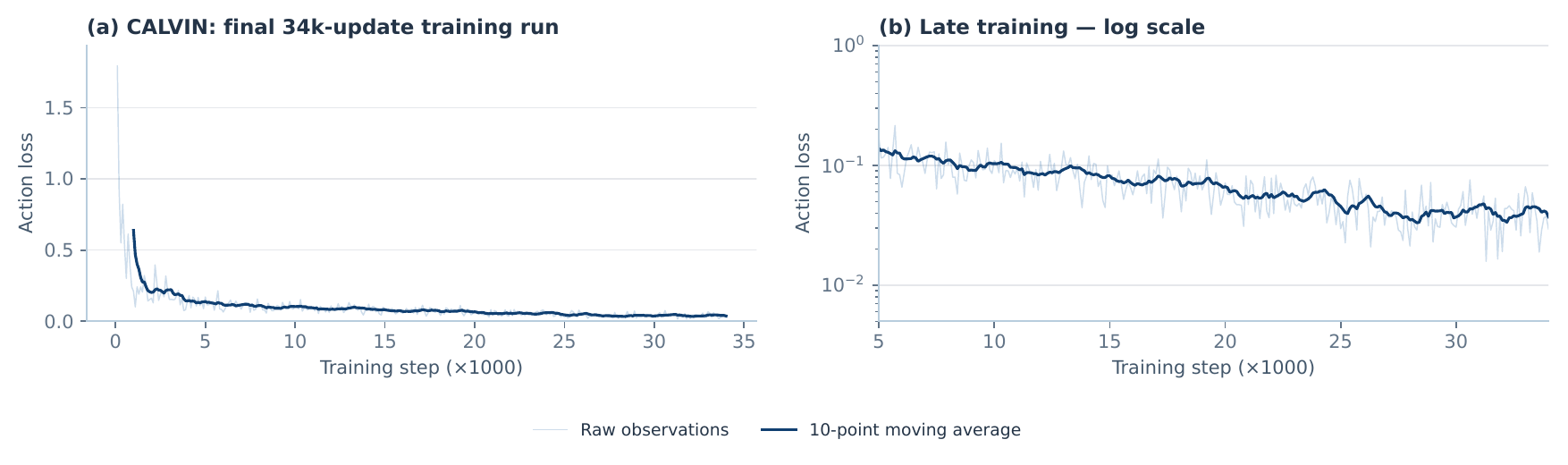}{1}{Action loss from the final CALVIN training run. All 340 observations at 100-update intervals are retained; the moving average uses ten observations. Nine earlier observations from preliminary runs in the same console file are excluded from this curve and accounted for in the audit. Final action loss is 0.02949.}{fig:calvin_loss}

Every row below uses the actual number of reached attempts. A task later in a sequence is observed only when its predecessors succeed. The table includes \texttt{close\_drawer} exactly once and contains 34 unique task types. Totals are 3,625 successes and 4,054 attempts; 26 task types achieve at least 85\% conditional success.
\begingroup\small
\renewcommand{\arraystretch}{1.06}
\begin{longtable}{ccc}
\caption{Complete CALVIN conditional task counts for the selected final evaluation.}\label{tab:calvin_tasks}\\
\toprule Task & Success / attempts & Rate (\%)\\\midrule\endfirsthead
\toprule Task & Success / attempts & Rate (\%)\\\midrule\endhead
\bottomrule\endfoot
close\_drawer & 182/182 & 100.00 \\
lift\_blue\_block\_drawer & 19/19 & 100.00 \\
lift\_blue\_block\_slider & 100/123 & 81.30 \\
lift\_blue\_block\_table & 143/143 & 100.00 \\
lift\_pink\_block\_drawer & 11/11 & 100.00 \\
lift\_pink\_block\_slider & 109/128 & 85.16 \\
lift\_pink\_block\_table & 120/129 & 93.02 \\
lift\_red\_block\_drawer & 17/17 & 100.00 \\
lift\_red\_block\_slider & 106/122 & 86.89 \\
lift\_red\_block\_table & 153/155 & 98.71 \\
move\_slider\_left & 231/233 & 99.14 \\
move\_slider\_right & 252/252 & 100.00 \\
open\_drawer & 307/320 & 95.94 \\
place\_in\_drawer & 147/151 & 97.35 \\
place\_in\_slider & 258/298 & 86.58 \\
push\_blue\_block\_left & 50/70 & 71.43 \\
push\_blue\_block\_right & 23/70 & 32.86 \\
push\_into\_drawer & 88/102 & 86.27 \\
push\_pink\_block\_left & 33/73 & 45.21 \\
push\_pink\_block\_right & 21/64 & 32.81 \\
push\_red\_block\_left & 50/74 & 67.57 \\
push\_red\_block\_right & 19/72 & 26.39 \\
rotate\_blue\_block\_left & 66/66 & 100.00 \\
rotate\_blue\_block\_right & 70/77 & 90.91 \\
rotate\_pink\_block\_left & 54/55 & 98.18 \\
rotate\_pink\_block\_right & 45/68 & 66.18 \\
rotate\_red\_block\_left & 60/61 & 98.36 \\
rotate\_red\_block\_right & 69/72 & 95.83 \\
stack\_block & 153/165 & 92.73 \\
turn\_off\_led & 138/150 & 92.00 \\
turn\_off\_lightbulb & 134/135 & 99.26 \\
turn\_on\_led & 168/168 & 100.00 \\
turn\_on\_lightbulb & 164/164 & 100.00 \\
unstack\_block & 65/65 & 100.00 \\
\end{longtable}\endgroup

\input{sections/extension_appendix.tex}

\end{document}

%% file: sections/authors.tex
\hypersetup{pdfauthor={Xin Ma, Furong Chen, Xinlong Man, Bo Zhang, Yalong Guo, Haolan Lei, Chaochen Cui, Hongyuan Zhang, Mulin Chen, Yi Wang, Lap-pui Chau, Chuang Yang}}
\newcommand{\roboseedauthorblock}{%
  \begin{minipage}{\textwidth}
  \centering
  {\normalsize
  Xin Ma\textsuperscript{1\textdagger},
  Furong Chen\textsuperscript{2\textdagger},
  Xinlong Man\textsuperscript{2\textdagger},
  Bo Zhang\textsuperscript{1},
  Yalong Guo\textsuperscript{3},
  Haolan Lei\textsuperscript{4}\\[2pt]
  Chaochen Cui\textsuperscript{2p},
  Hongyuan Zhang\textsuperscript{5},
  Mulin Chen\textsuperscript{1},
  Yi Wang\textsuperscript{6},
  Lap-pui Chau\textsuperscript{6},
  Chuang Yang\textsuperscript{6*p}\par}
  \vspace{4pt}
  {\small
  \textsuperscript{1}Northwestern Polytechnical University\par
  \textsuperscript{2}Jiangsu Yunmu Intelligent Manufacturing Co., Ltd.\par
  \textsuperscript{3}National University of Singapore\par
  \textsuperscript{4}Xi'an University of Posts and Telecommunications\par
  \textsuperscript{5}The University of Hong Kong\par
  \textsuperscript{6}The Hong Kong Polytechnic University\par}
  \end{minipage}%
}
\author{\roboseedauthorblock}

\newcommand{\roboseedauthornotes}{%
  \par\vfill
  \begingroup\footnotesize
  \noindent\rule{0.3\linewidth}{0.4pt}\par
  \vspace{2pt}
  \noindent\textsuperscript{\textdagger}These authors contributed equally to this work.\par
  \noindent\textsuperscript{*}Corresponding author.\quad
  \textsuperscript{p}Project Leader.\par
  \noindent Email:
  \href{mailto:omtcyang@gmail.com}{\nolinkurl{omtcyang@gmail.com}}\par
  \endgroup
}

%% file: sections/reproductions.tex
\input{sections/vla_extensions.tex}
\input{sections/wam_reproductions.tex}

%% file: sections/vla_extensions.tex
\subsection{LingBot-VLA 2.0}\label{sec:lingbot}
\subsubsection{Mechanism and Reproduction Boundary}
LingBot-VLA 2.0 combines a Qwen3-VL-4B prefix with sparse action experts and predictive visual supervision\citep{lingbot2}. The inspected policy uses 32 routed experts, top-4 routing and a shared expert in the action stream. Depth and video teachers supervise current and future queries. These mechanisms target control specialization and predictive representations; their benefits depend on the optimization and data regime. A sparse action expert does not make the dense VLM, camera encoding, communication, or teacher supervision free.

\fig{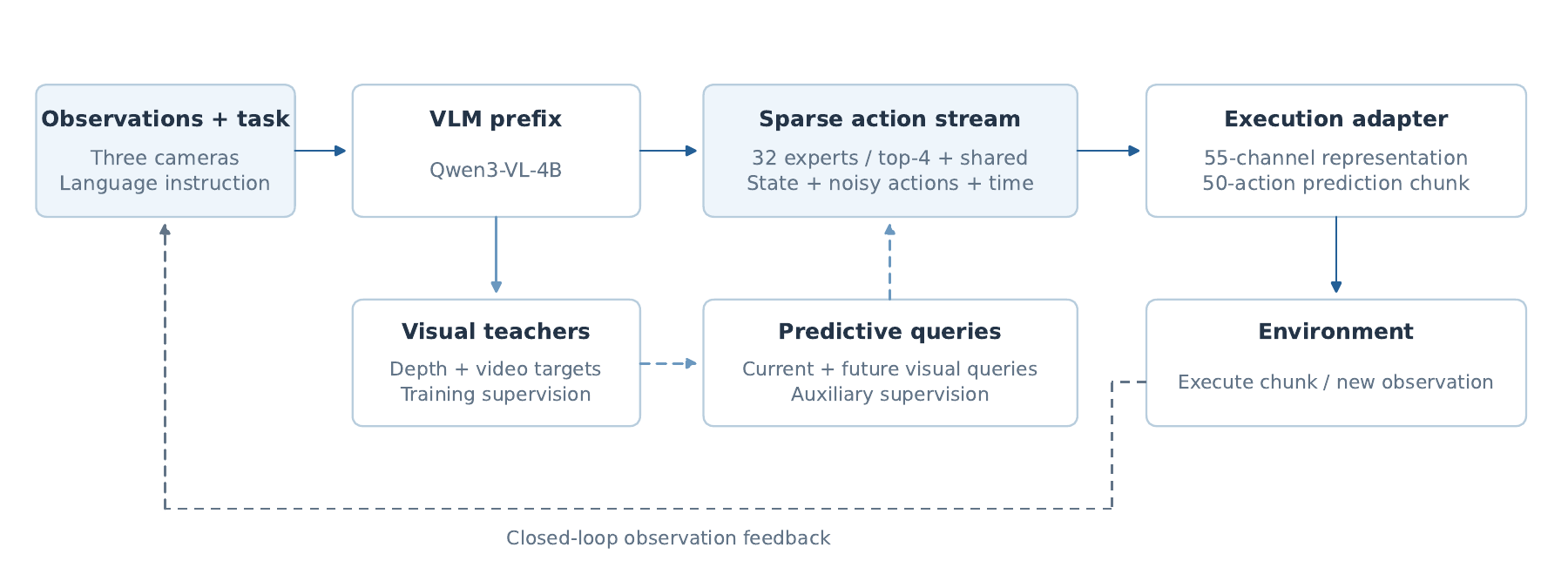}{1}{LingBot-VLA 2.0 reproduction interfaces. Multi-view images and language enter Qwen3-VL-4B. The action stream uses 32 routed experts with top-4 routing and a shared expert. Depth and video teachers supervise predictive queries during training; dashed arrows denote training supervision or environment feedback.}{fig:lingbot_architecture}

\subsubsection{Experimental Setup}
The local RoboTwin recipe uses Muon, three training GPUs, micro-batch 32, global batch 96, 50k updates, and warmup ratio 0.0333. Its saved policy specifies 55 action/state channels, three-camera input, chunk length 50, ten inference steps, and BF16 evaluation with compilation. The inspected configuration has \texttt{bias\_update\_speed=0}, sequence-wise routing loss $10^{-3}$ and router z-loss $10^{-4}$. Thus this run must not be described as validating a loss-free router update. Teacher dependencies and normalization conversion are part of the training burden. Both the released post-trained checkpoint and the locally post-trained checkpoint are evaluated; the foundation pre-train itself is inherited, not reproduced.

\begin{table}[!htbp]\centering\small
\caption{LingBot-VLA 2.0 reproduction settings. The local run changes the adaptation budget; it does not repeat the foundation pre-train.}\label{tab:ling_setup}
\begin{tabularx}{\linewidth}{C{4.0cm}Y}\toprule
Setting & Observed configuration\\\midrule
Initialization / backbone & Released LingBot foundation / Qwen3-VL-4B plus sparse action experts\\
Input / native representation & Three cameras; 55 state/action channels\\
Optimizer / local batch & Muon; micro-batch 32, global batch 96\\
Adaptation / training GPUs & 50,000 updates; three H200 GPUs\\
Warmup / terminal base LR & Ratio 0.0333 / $5\times10^{-5}$\\
Action chunk / integration & 50 predicted/executed actions; ten flow steps\\
Completed evaluation & Clean; unseen instructions; seed 0; BF16 and compilation\\
Coverage / rollout slots & 50 tasks $\times$ 100 episodes; four GPUs $\times$ three slots\\\bottomrule
\end{tabularx}\end{table}

\subsubsection{RoboTwin Success Results}
The two original \texttt{stats.txt} files each contain exactly 50 completed tasks, 100 episodes per task, and zero skipped tasks. The released checkpoint achieves $4486/5000=89.72\%$ Clean success; the local checkpoint achieves $2889/5000=57.78\%$. Randomized evaluation is absent from these completed records. Exact counts take precedence over the rounded 89.7\% and 57.8\% in the source summaries.

\begin{table}[!htbp]\centering\small
\caption{LingBot-VLA 2.0 closed-loop results on its common local evaluator. Missing Randomized outcomes remain unevaluated.}\label{tab:ling_results}
\begin{tabularx}{\linewidth}{C{4.0cm}YY}\toprule
Metric & Official weights & Local 50k checkpoint\\\midrule
Clean successes / episodes & 4486/5000 & 2889/5000\\
Clean success rate & 89.72\% & 57.78\%\\
Randomized success rate & -- & --\\
Completed / skipped tasks & 50 / 0 & 50 / 0\\
Local minus official & \multicolumn{2}{c}{$-31.94$ percentage points on Clean}\\\bottomrule
\end{tabularx}\end{table}

\subsubsection{Training Resources and Rollout Runtime}

The final training progress interval is 58:07:18 on three GPUs, giving \textbf{174.365 H200 GPU-hours}. This excludes setup and the final asynchronous export; it is a scoped training interval, not the total cost of developing the reproduction. The final HF export completes successfully. The six weight shards occupy \textbf{25.504 decimal GB} for either checkpoint. This is the saved export size, not BF16 resident inference memory. The local evaluation takes 32,130 seconds (8.925 h) with four GPUs and 12 slots; the official-checkpoint evaluation takes 27,660 seconds (7.683 h). These exact totals replace a source note that loosely described both runs as about 8.9 h.

Local sample exposure is $96\times50{,}000=4.8$ million windows. The source reference recipe reports batch 1024 and 50k updates, or 51.2 million windows: a factor of 10.67 more. At batch 96, matching that count would require about 533,334 updates. This arithmetic is a budget conversion, \emph{not} a prediction of the resulting score. Larger batch, learning-rate warmup, data conversion, teacher paths, precision, and initialization can also affect the result. The comparison is consistent with insufficient adaptation, but does not isolate exposure as the sole cause. The three observations from LingBot and ABot do not constitute a controlled exposure--performance scaling curve.

\begin{table}[!htbp]\centering\small
\caption{LingBot resource observations. Training time is the final progress interval; evaluation time includes the complete task queue. Artifact GB do not measure resident BF16 VRAM.}\label{tab:ling_cost}
\begin{tabularx}{\linewidth}{C{4.0cm}YY}\toprule
Metric & Official weights & Local 50k checkpoint\\\midrule
Reference / local sample exposure & 51.2M (source recipe) & 4.8M\\
Measured adaptation wall time & -- & 58:07:18\\
Measured H200 training GPU-hours & -- & 174.365\\
Clean evaluation wall time & 7.683 h & 8.925 h\\
Clean evaluation GPU-hours & 30.733 & 35.700\\
Primary weight artifact & 25.504 GB & 25.504 GB\\
Inference peak VRAM / p95 latency & -- / -- & -- / --\\\bottomrule
\end{tabularx}\end{table}

\subsubsection{Training Convergence}
The new log extraction retains all 50,000 update-level records. Logged total loss falls from 0.3906 at update one to 0.0194 at update 50,000, while the VLA component changes from 0.3594 to 0.0149. The depth/video losses and learning-rate series are retained in the data CSV. These optimization observations establish learning progress under the native objective; the independently measured 57.78\% Clean result determines the checkpoint's task capability. Neither low final loss nor the local loss slope predicts recovery to the official success rate.

\begin{figure}[!htb]\centering
\includegraphics[width=\linewidth]{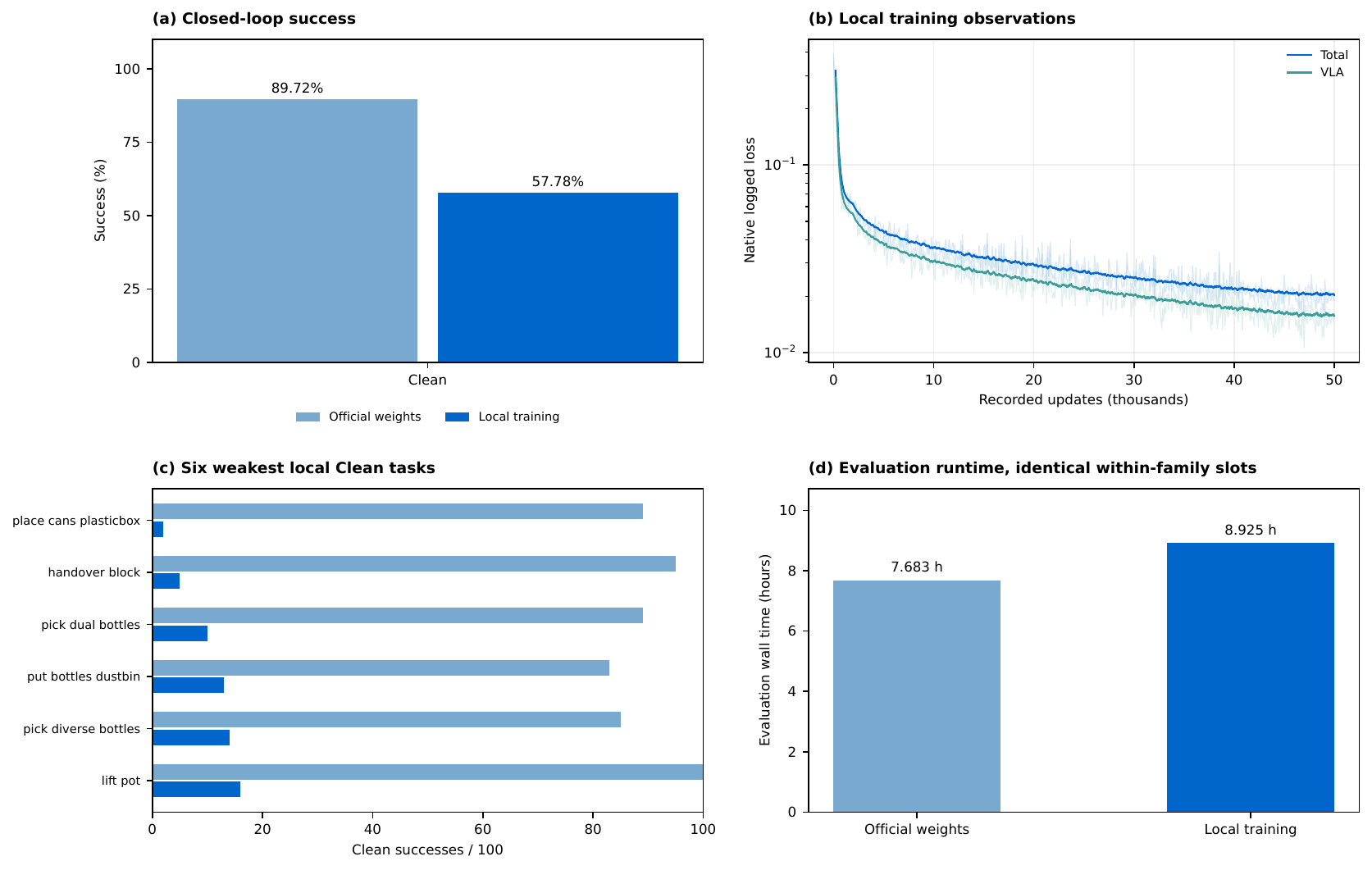}
\caption{LingBot-VLA 2.0 experimental results. (a) Official and local Clean outcomes use 5,000 episodes each. (b) Total and VLA loss come from all 50,000 update records; light traces display every 83rd record and dark traces use a 200-observation moving average. (c) The six lowest local Clean tasks are compared with the official checkpoint under the same within-family evaluator. (d) Full evaluation queue time uses the same four-GPU, 12-slot configuration. Randomized outcomes and serving latency are unavailable.}\label{fig:ling_experiments}
\end{figure}

\subsubsection{Task-Level Profiles and Failure Diagnosis}
The largest local failures include plastic-box can placement (2/100), block handover (5/100), and dual-bottle picking (10/100), while roller grasping reaches 100/100. Exact task identifiers and all counts appear in Appendix~\ref{app:extension}. This unevenness is useful for selecting diagnostic tasks. It does not prove a particular teacher, camera layout, or MoE component caused a gain or failure. The 31.94-point official-to-local Clean gap establishes substantial recipe sensitivity under the two available checkpoints. A matched exposure study and teacher/MoE ablations are needed to quantify mechanism value. The current result validates the local training-to-export-to-rollout chain, and provides a checkpoint to improve; it does not reproduce the official checkpoint's capability level.

\begin{table}[H]\centering\small
\caption{Ten lowest local LingBot Clean task results. Each count is out of 100; the difference is local minus official in percentage points. All 50 tasks remain in Appendix~\ref{app:extension}.}\label{tab:ling_tasks}
\begin{tabularx}{\linewidth}{C{6.0cm}YYY}\toprule
Task & Official & Local & Difference\\\midrule
\input{table_ling_weak_tasks.tex} 
\bottomrule\end{tabularx}\end{table}

%% file: table_ling_weak_tasks.tex
place\_cans\_plasticbox & 89/100 & 2/100 & -87\\
handover\_block & 95/100 & 5/100 & -90\\
pick\_dual\_bottles & 89/100 & 10/100 & -79\\
put\_bottles\_dustbin & 83/100 & 13/100 & -70\\
pick\_diverse\_bottles & 85/100 & 14/100 & -71\\
lift\_pot & 100/100 & 16/100 & -84\\
move\_can\_pot & 88/100 & 16/100 & -72\\
scan\_object & 96/100 & 16/100 & -80\\
click\_bell & 58/100 & 18/100 & -40\\
move\_stapler\_pad & 70/100 & 33/100 & -37\\

%% file: sections/wam_reproductions.tex
\section{Empirical Evaluation of World--Action Models}\label{sec:wam}
This chapter supplies the WAM evidence for the proposed evaluation framework, examining how predictive representations, inherited priors and task conditions affect capability and resource interpretation. OpenWAM official/local comparisons measure adaptation outcomes and costs; LiLa-WAM tests the practical scope of compact latent prediction with visual task selection. Implementation inspection identifies which mechanisms are executable, and the training and rollout audits establish which research-value and reproduction-evidence criteria are satisfied. Each method's setup, results and diagnosis feeds the dimension profiles in Chapter~\ref{sec:assessment}.

\paragraph{World--action modeling.}
A world--action model (WAM) learns representations of future observations together with a robot action policy. Given the current observation $o_t$, task condition $c$ and robot state $q_t$ when available, its coupled prediction can be written schematically as
\begin{equation}
(\hat z_{t+1:t+K},\hat A_{t:t+H-1})=F_\theta(o_t,c,q_t),
\end{equation}
where $z$ denotes a future visual representation and $A$ an action chunk. This notation summarizes the prediction targets; it does not require both outputs to be decoded at deployment. Future representations can retain video structure, as in OpenWAM, or occupy a compact feature space, as in LiLa-WAM. Predictive supervision aims to connect visual change with controllable motion. Its contribution must be tested by controlled ablation rather than inferred from a success score alone.

\paragraph{Shared reproduction protocol.}
For each WAM, we identify the inherited visual or video prior, frozen modules, task condition, prediction targets, action representation and execution adapter. We then report local adaptation settings, checkpoint-specific rollout outcomes and measured resource costs. OpenWAM uses text conditioning and an end-effector interface; the inspected LiLa-WAM uses task-name-selected visual transition tokens and joint actions. These differences remain explicit in comparisons. The two methods retain Sections 4.1 and 4.2, respectively.

\subsection{OpenWAM}\label{sec:openwam}
\subsubsection{World--Action Coupling and Interfaces}
OpenWAM develops a modular world--action modeling stack with dedicated world and action capacity and explicit interaction between them\citep{openwam}. The inspected run combines Wan2.2-TI2V-5B, Wan VAE and ActionDiT in a dual-system joint-self-attention architecture. VAE and T5 text encoding are frozen. Training uses three views, 33-frame windows with video stride four, and configured height/width $384\times320$. Twenty physical end-effector action channels map to slots 0--9 and 34--43 in an 80-channel unified representation. This differs materially from ABot's joint-action interface. Inverse kinematics and the execution adapter are therefore part of the comparison, as are the visual inputs and foundation initialization.

\fig{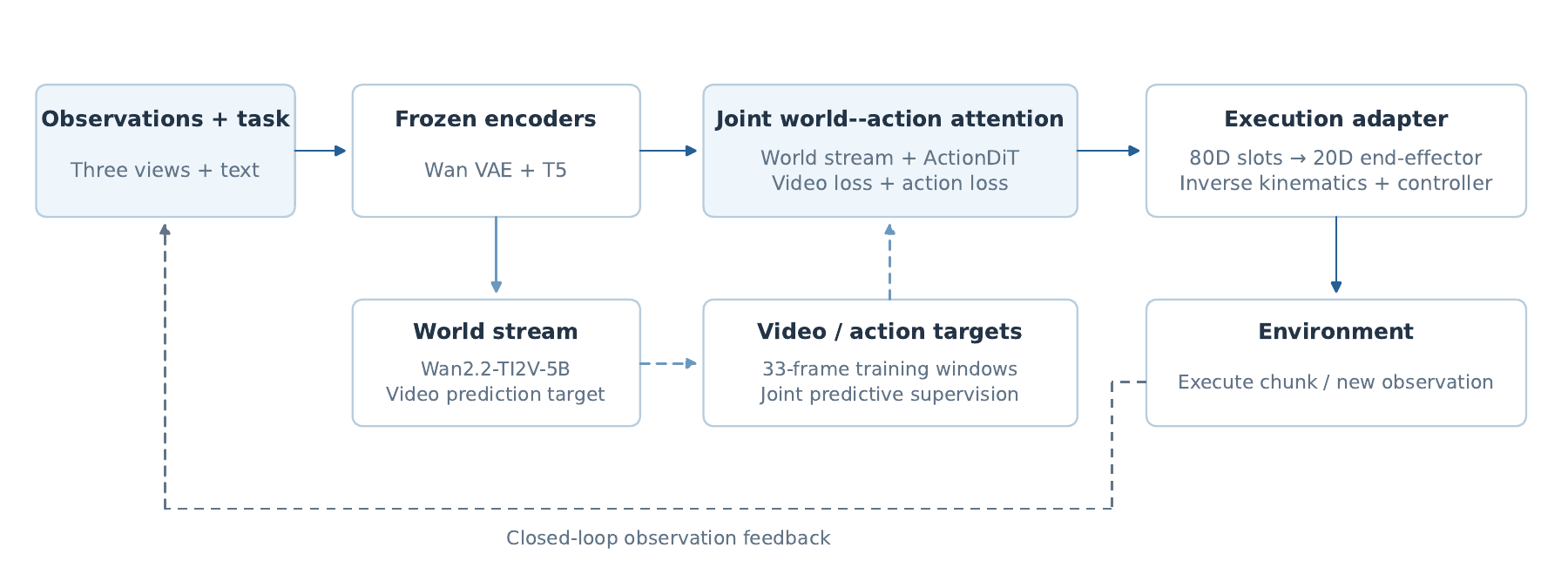}{1}{OpenWAM reproduction interfaces. Frozen Wan VAE and T5 encoding condition a Wan2.2 world stream and ActionDiT action stream coupled by joint self-attention. Training supervises video and action targets; the execution adapter selects 20 physical end-effector channels from the 80-channel representation.}{fig:openwam_architecture}

\subsubsection{Experimental Setup}
The local run starts from the released Foundation checkpoint (step 154000), rather than training the video prior or robotic foundation from scratch. It uses four H200 NVL GPUs, BF16, ZeRO-2, AdamW, global batch 64, 80k updates, learning rate $10^{-4}$, 5\% warmup, and equally weighted video/action losses. The saved \texttt{max\_steps=80000} and the final checkpoint establish the stopping point; the \texttt{num\_epochs=1} field alone does not describe the effective training duration. The final reported total/video/action losses are 0.08358/0.08310/0.00048. Their numerical magnitudes cannot be compared with ABot's or LingBot's losses because targets, normalization and loss definitions differ.

\begin{table}[!htbp]\centering\small
\caption{OpenWAM local adaptation and evaluation settings. The released Foundation initialization is inherited.}\label{tab:openwam_setup}
\begin{tabularx}{\linewidth}{C{4.2cm}Y}\toprule
Setting & Observed configuration\\\midrule
World / action modules & Wan2.2-TI2V-5B plus Wan VAE / ActionDiT\\
Training views / video window & Three views; 33 frames, stride four; $384\times320$\\
Native action / unified channels & 20D end-effector commands / 80D slots\\
Updates / local global batch & 80,000 / 64 (four GPUs $\times$ 16)\\
Optimizer / learning-rate schedule & AdamW; $10^{-4}$, cosine, warmup ratio 0.05\\
Training precision / parallelism & BF16 / ZeRO-2 on four H200 NVL GPUs\\
Evaluation scenes / instructions & Clean and Randomized / unseen, seed 0\\
Coverage per scene / slots & 50 tasks $\times$ 100 episodes / four GPUs $\times$ four slots\\
Observation and success checks & Observation at chunk start; success checked after every action\\\bottomrule
\end{tabularx}\end{table}

\subsubsection{RoboTwin Success Results}
Two checkpoints each have 100 original task/scene result files, covering all 50 tasks in both scenes. The reported rates are exact multiples of 1/100; reconstructed successes sum to the source totals in Table~\ref{tab:openwam_results}.

\begin{table}[!htbp]\centering\small
\caption{OpenWAM closed-loop outcomes. Official and local checkpoints share the adapted OpenWAM evaluator and each cover 10,000 episodes.}\label{tab:openwam_results}
\begin{tabularx}{\linewidth}{C{4.2cm}YY}\toprule
Metric & Official weights & Local BS64-80K\\\midrule
Clean successes / episodes & 4701/5000 & 4565/5000\\
Clean success rate & 94.02\% & 91.30\%\\
Randomized successes / episodes & 4683/5000 & 4570/5000\\
Randomized success rate & 93.66\% & 91.40\%\\
Two-scene mean & 93.84\% & 91.35\%\\
Local minus official & \multicolumn{2}{c}{$-2.49$ percentage points overall}\\
\bottomrule\end{tabularx}\end{table}

The released OpenWAM checkpoint achieves 93.84\% across 10k episodes. The local BS64-80K checkpoint reaches 91.35\%, with 91.30\% Clean and 91.40\% Randomized success. The matched local evaluator gives a \textbf{2.49-point adaptation gap}. Local exposure is 5.12 million video windows, versus a source reference estimate of $118655\times256=30.376$ million, or 16.86\%. That estimate combines the released update count with the reported reference global batch; it does not establish the official GPU count or monetary cost. A video window and an ABot training example contain different information and compute, so these counts are not interchangeable units of sample efficiency.

\subsubsection{Training Resources and Rollout Runtime}
The recorded 65:31:01 local training interval gives \textbf{262.068 H200 GPU-hours}; the saved checkpoint is \textbf{24.814 GB}. The local evaluator requires 17:03:02 for Clean and 16:42:39 for Randomized (33:45:41 total). Four allocated GPUs imply 135.046 evaluation GPU-hours, roughly 52\% of the recorded downstream training budget. The corresponding official-weight evaluation takes 29:10:05 (116.672 GPU-hours). Identical model structure with different wall time illustrates why rollout duration cannot be treated as pure neural-network latency. Success-dependent episode length, CPU load, rendering and contention can all change runtime.

\begin{table}[!htbp]\centering\small
\caption{OpenWAM resource observations. Downstream training excludes Foundation pre-train; evaluation-device hours assume the four recorded GPUs are allocated throughout each queue.}\label{tab:openwam_cost}
\begin{tabularx}{\linewidth}{C{4.2cm}YY}\toprule
Metric & Official weights & Local BS64-80K\\\midrule
Reference / local sample exposure & 30.376M (source recipe) & 5.120M\\
Measured adaptation wall time & -- & 65:31:01\\
Measured training GPU-hours & -- & 262.068\\
Clean evaluation wall time & 14:04:28 & 17:03:02\\
Randomized evaluation wall time & 15:05:37 & 16:42:39\\
Total evaluation GPU-hours & 116.672 & 135.046\\
Primary weight artifact & 24.814 GB & 24.814 GB\\
Inference peak VRAM / p95 latency & -- / -- & -- / --\\\bottomrule
\end{tabularx}\end{table}

Both OpenWAM evaluations use commit \texttt{0aeea2d} and an adapted execution path: observation and policy requests occur at the first step of an action chunk, while success and step limits are checked after every executed action. Four independent services per GPU provide 16 slots across four GPUs. The two OpenWAM checkpoints share this protocol, supporting their internal comparison. This does not establish equivalence to the upstream evaluator or to ABot's recorded commit \texttt{13c3c47}. The small Clean--Randomized difference is evidence about the tested randomized distribution, which is included in adaptation data; it is not a clean-only-to-randomized generalization test.

\subsubsection{Training Convergence}
The console supplies repeated rank/progress-bar records. For an auditable display we retain the first occurrence at each positive progress counter, yielding 80,000 observations. The displayed total/action losses change from 0.2679/0.0537 near the start to 0.0836/0.0005 at the end; video loss remains the dominant numerical contribution. These are four-decimal console values. The more precise final tracker values, 0.08358/0.08310/0.00048 for total/video/action, are retained separately and are not substituted throughout the curve. The progress display can lead or lag the associated optimizer update by one because the logger sets its postfix before advancing its counter; this precision is sufficient to visualize convergence, not to infer a particular update's causal effect.

\fig{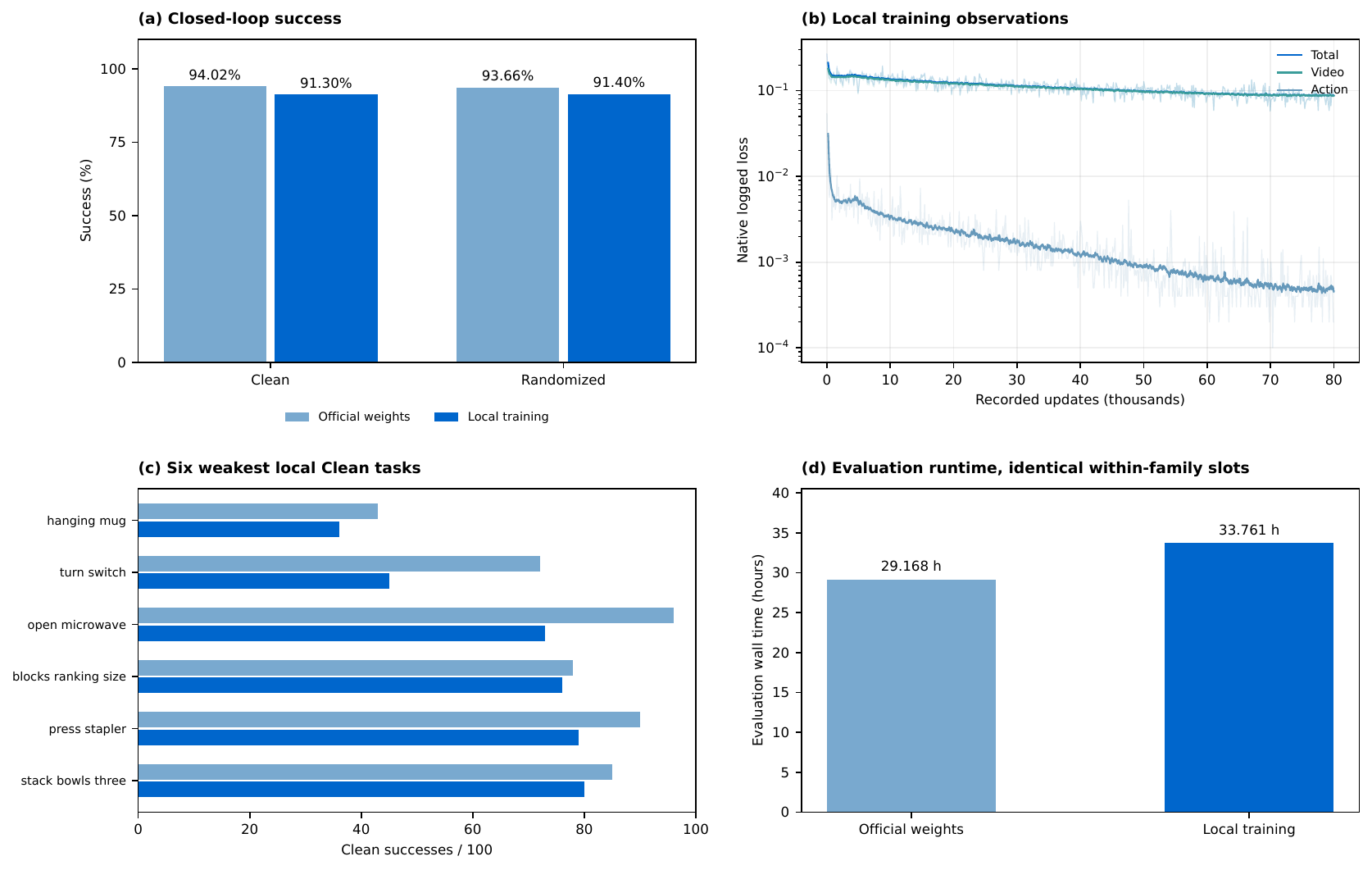}{1}{OpenWAM experimental results. (a) Separate Clean and Randomized rates use 5,000 episodes per scene and checkpoint. (b) Native total, video and action loss use 80,000 selected console observations, every 133rd raw value, and a 200-observation moving average; console losses are rounded and progress counters have a possible one-update offset. (c) Six weakest local Clean tasks retain the same within-family checkpoint comparison. (d) Evaluation wall time covers both scenes on four GPUs and 16 slots; it is not policy inference latency.}{fig:openwam_experiments}

\subsubsection{Task-Level Profiles and Failure Diagnosis}
Table~\ref{tab:openwam_tasks} follows the ten lowest local Clean task rates and retains both scenes. Local Clean and Randomized averages are nearly equal, but the task-specific gaps differ; an overall mean can hide substantial task deficits. For instance, the local \texttt{turn\_switch} results are 45/100 and 57/100, while \texttt{hanging\_mug} falls to 36/100 and 34/100. Their weakness is observed in both scene distributions rather than confined to randomized rendering.

\begin{table}[!htbp]\centering\small
\caption{Ten lowest local OpenWAM Clean tasks. Each C/R entry is a pair of success counts out of 100 per scene; differences are local minus official two-scene means in percentage points.}\label{tab:openwam_tasks}
\begin{tabularx}{\linewidth}{C{5.4cm}YYY}\toprule
Task & Official C/R & Local C/R & Difference\\\midrule
\input{table_openwam_weak_tasks.tex} 
\bottomrule\end{tabularx}\end{table}

\subsubsection{Algorithmic and Practical Value}
OpenWAM supplies the highest observed two-scene rate among the locally trained methods in this snapshot. Relative to the ABot Qwen3.5-4B record, its local rate is 5.85 points higher and its recorded downstream GPU-hours are about 33.5\% lower. These are useful procurement observations under the available recipes, but do not isolate an architectural advantage: initialization, data windows, observations, action spaces, exposure, kernels and evaluator paths differ. Training a video foundation is outside the reported 262.068 GPU-hours, and no full-lifecycle cost advantage has been demonstrated.

Within OpenWAM, the largest average local deficits occur on \texttt{turn\_switch} ($-17.5$ points), \texttt{hanging\_mug} ($-13.5$), \texttt{open\_microwave} ($-12.5$) and \texttt{move\_stapler\_pad} ($-11.0$). These deficits occur across both domains and motivate investigating contact and alignment behavior. The available experiment changes an entire training recipe; it does not test whether explicit future prediction, joint attention or the video loss individually improves control. Such claims remain architectural hypotheses locally, even when supported by the original authors' experiments.

\subsubsection{Additional Video-Backbone Training Prefix}
The October 28 inspection also finds an OpenWAM Wan2.1-VACE-1.3B run configured for 80k updates, global batch 64 and the same 80-channel action representation. The latest saved trainer state identifies update 50,000; the console reaches displayed progress 51,180/80,000. The saved 50k artifact is 17.211 GB. Its final displayed total/video/action losses are 0.0284/0.0275/0.0009, with a displayed interval around 71:04:28. These values describe an unfinished training prefix and native rounded logging, not a complete 80k adaptation cost. No corresponding task-result set was found, so this candidate receives no capability score and is recorded outside the 12 evaluated checkpoint profiles. Its source configuration and trainer state are retained in the refresh audit.

\subsection{LiLa-WAM}\label{sec:lila}
\subsubsection{Compact Latent Prediction and Task Specification}
LiLa-WAM couples future-feature prediction and action generation using a frozen DINOv3 ViT-L/16 and a compact action model\citep{lila}. The source describes about 0.5B total and 0.2B trainable parameters. These are source-reported scale descriptors, not a local enumeration of all resident tensors. The inspected setting uses a head image at $320\times240$, 14 joint-action channels, 32 predicted actions, execution of the first 16, ten flow-matching steps, and spline smoothing.

\textbf{The policy is task-conditioned through visual tokens.} The inspected \texttt{set\_task(task\_name)} loads a precomputed \texttt{task\_cond.npy}; \texttt{\_predict\_chunk} explicitly leaves its instruction argument unused. Consequently, an evaluator labeled \texttt{unseen} does not establish understanding of unseen natural-language instructions in this run. VTT replaces language with a known task's visual transition vector. This can reduce runtime and training burden for a fixed task inventory, but requires constructing and distributing task conditions and does not provide an instruction interface for arbitrary new requests. Its operational value must be judged against that requirement.

\fig{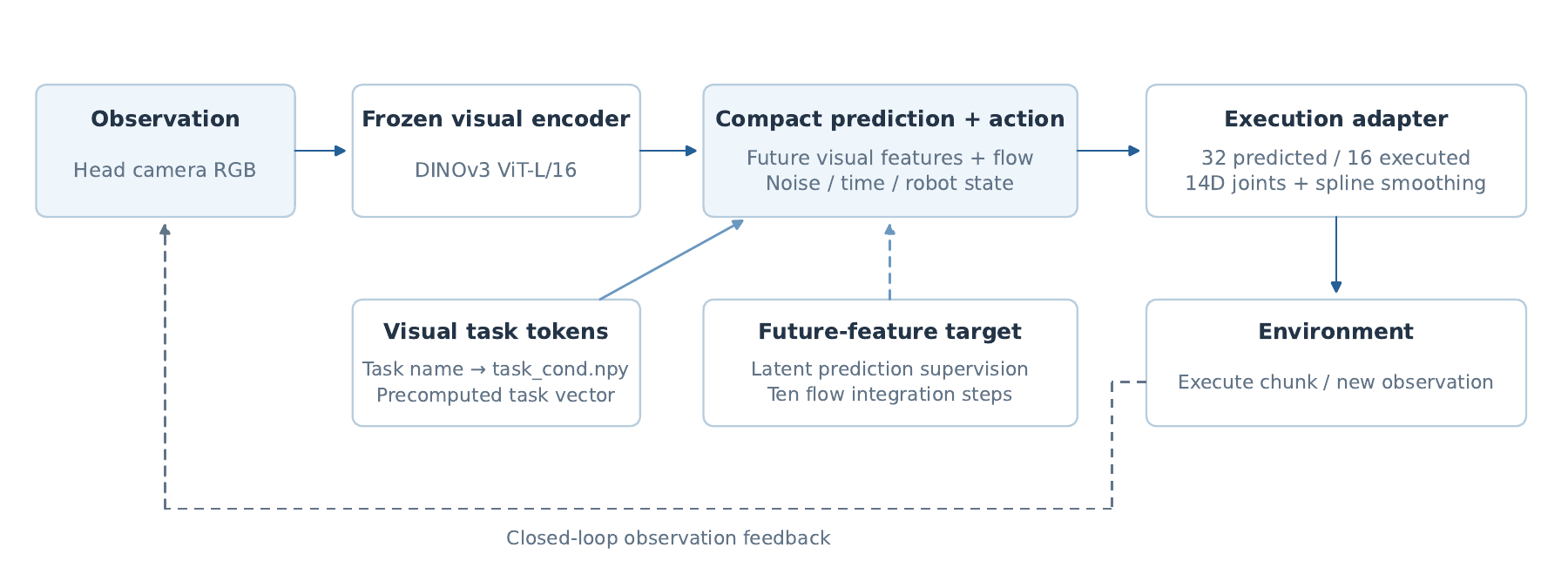}{1}{LiLa-WAM reproduction interfaces. A frozen DINOv3 encoder provides head-camera features. A task name selects precomputed visual transition tokens; the inspected policy does not use the instruction argument. Compact future-feature prediction and action generation share the task condition. The controller executes 16 of 32 predicted actions with spline smoothing.}{fig:lila_architecture}

\subsubsection{Experimental Setup}
\begin{table}[!htbp]\centering\small
\caption{LiLa-WAM settings shared by the official-weight and local stage-2 evaluations. Its visual task interface is explicit.}\label{tab:lila_setup}
\begin{tabularx}{\linewidth}{C{4.2cm}Y}\toprule
Setting & Observed configuration\\\midrule
Visual encoder / policy scale & Frozen DINOv3 ViT-L/16 / approximately 0.5B (source)\\
Task specification & Task-name-selected precomputed visual transition vector\\
Input view / size & Head camera / $320\times240$\\
Action / internal state channels & 14 / 16\\
Prediction / execution horizon & 32 / 16 actions\\
Integration / smoothing & Ten flow steps / B-spline\\
Completed checkpoints & Official stage-2 weights and local stage-2 epoch four\\
Evaluation scenes / coverage & Clean / 50 tasks $\times$ five episodes\\
Hardware / slots & One H200 NVL / four parallel workers\\\bottomrule
\end{tabularx}\end{table}

\subsubsection{RoboTwin Success Results}
The official weights achieve $224/250=89.60\%$ Clean success; the local stage-2 epoch-4 weights achieve $222/250=88.80\%$. Each evaluation covers the same 50 task names with five episodes per task and four workers. All worker JSON counts match their per-task final log counters. The local evaluation setting resolves to the saved stage-2 checkpoint from October 23. The official results are retained in a separate backup directory because the active worker-result filenames were reused by local evaluation.

The local-minus-official difference is $-0.80$ percentage points, or two successes across 250 episodes. Nineteen tasks change their success count; 31 are unchanged. Thirty-two official and 33 local tasks achieve 5/5, while seven official and five local tasks fall below 4/5. Aggregate similarity can therefore coexist with task-level changes. Five episodes give coarse 20-point per-task increments; these observations do not prove statistical equivalence or isolate the latent-prediction mechanism. Randomized results remain unavailable for both checkpoints.

\begin{table}[!htbp]\centering\small
\caption{LiLa-WAM official/local Clean results. Each column represents a complete 250-episode evaluation, with task-name-selected visual conditions.}\label{tab:lila_results}
\begin{tabularx}{\linewidth}{C{4.2cm}YY}\toprule
Metric & Official weights & Local stage-2 epoch 4\\\midrule
Clean successes / episodes & 224/250 & 222/250\\
Clean success rate & 89.60\% & 88.80\%\\
Working Wilson 95\% interval & 85.20\%--92.80\% & 84.29\%--92.14\%\\
Randomized success rate & -- & --\\
Tasks with five successes & 32/50 & 33/50\\
Tasks below four successes & 7/50 & 5/50\\
Local minus official & \multicolumn{2}{c}{$-0.80$ percentage points}\\\bottomrule
\end{tabularx}\end{table}

The working pooled Wilson intervals describe episode sampling under a fixed tested task mixture. They exclude training-seed variation, task-population uncertainty and protocol bias. Overlapping intervals and proximity to the source's 90.48\% are not equivalence tests. The completed local evaluation establishes that the saved local training prefix can execute this task protocol; the contribution of individual components still requires controlled ablation.

\subsubsection{Local Training Convergence and Saved Prefixes}

Stage-1 checkpoints extend through epoch 12 and stage-2 checkpoints through epoch 4. The October 26 local rollout now verifies the epoch-4 stage-2 checkpoint, superseding the previous pending status. The stage-2 training log enters epoch 5; successful evaluation of epoch 4 does not establish completion of the configured 40-epoch schedule. The saved learning rate is $4\times10^{-5}$ while \texttt{lr\_min} is $5\times10^{-5}$, and the log warns that this schedule increases rather than decreases the rate. The observed 88.80\% is tied to that actual recipe; the effect of correcting the schedule remains untested.

Loss extraction is restricted to the completed saved prefixes: 23,643 observations over stage-1 epochs 1--12 and 7,881 over stage-2 epochs 1--4. Seven observations from partial stage-1 epoch 13 and four from partial stage-2 epoch 5 are excluded and counted in the audit. The update axis is reconstructed from the epoch, batch index and 39,406 minibatches per epoch; counters restart for stage 2. Stage-1 logged action MSE changes from 0.3473 to 0.0014; stage 2 starts at 0.0019 and its last selected record is 0.0007. These are component losses at sampled updates. The epoch average is the full training objective, which also includes feature supervision; it is not the same scalar as action MSE.

\begin{table}[!htbp]\centering\small
\caption{LiLa local training prefixes. Summed epoch timers exclude startup, scans, checkpoint export and partial epochs. Local rollout is now verified, but the epoch timers lack an independent device-allocation trace and are not inserted into the H200 cost score.}\label{tab:lila_training}
\begin{tabularx}{\linewidth}{C{4.0cm}YYY}\toprule
Training prefix & Saved epochs & Epoch timers (h) & Last epoch total loss\\\midrule
\input{table_lila_training_prefix.tex} 
\bottomrule\end{tabularx}\end{table}

\fig{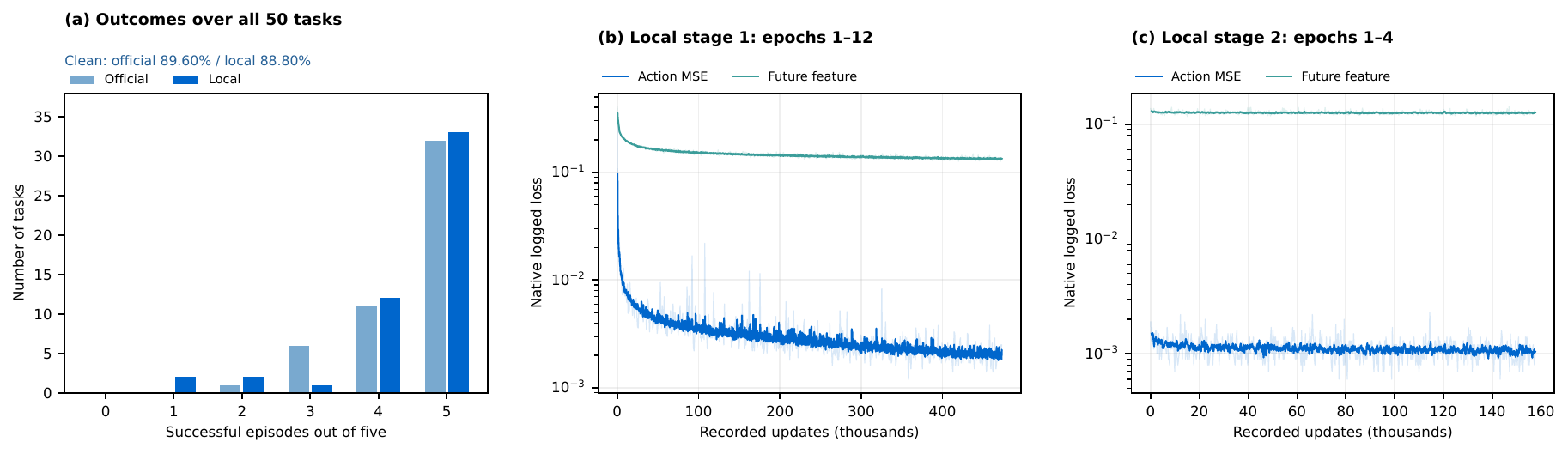}{1}{LiLa-WAM experimental results. (a) Official/local task-outcome distributions over 50 tasks with five episodes each; compact text reports Clean success, 89.60\%/88.80\%. (b,c) Local action-MSE and future-feature curves cover saved stage-1 epochs 1--12 and stage-2 epochs 1--4, with separate update axes. Light curves subsample every 39th/13th observation; dark curves use 20-observation moving averages. The local rollout uses the stage-2 epoch-4 checkpoint reached by these training prefixes.}{fig:lila_experiments}

\subsubsection{Training and Deployment Resource Boundaries}

The published single-24-GB-GPU feasibility and approximately 110 RTX-5090 GPU-hours in the repository are \emph{source claims}, not locally measured H200 training costs. They are therefore not entered into the local cost score. The official action checkpoint is loaded separately from DINOv3, and local training checkpoints can contain optimizer state. A single checkpoint's file size would undercount required dependencies or count training-only tensors; no deployment-package score is assigned until a clean inference bundle is measured. One H200 with four workers proves feasibility on that machine, not minimum VRAM or a robot's real-time deadline.

\subsubsection{Task-Level Profiles and Evidence Limits}
\begin{table}[!htbp]\centering\small
\caption{Ten lowest local LiLa-WAM Clean task results, retaining the official-weight comparator. A single extra success changes a task's displayed rate by 20 points; these rows should not be treated as precise task rankings.}\label{tab:lila_tasks}
\begin{tabularx}{\linewidth}{C{5.2cm}YYY}\toprule
Task & Official / 5 & Local / 5 & Gap (pp)\\\midrule
\input{table_lila_weak_tasks.tex} 
\bottomrule\end{tabularx}\end{table}

The lowest local tasks include \texttt{place\_a2b\_left} and \texttt{place\_dual\_shoes} at 1/5, and \texttt{open\_microwave} at 2/5. Conversely, \texttt{stamp\_seal} rises from 2/5 with official weights to 5/5 locally. Such changes motivate larger matched evaluations and failure replay, rather than a precise task ranking from five episodes. High success on known task vectors does not demonstrate free-form instruction understanding. The next extension is to increase identical scene sampling for both checkpoints and add Randomized coverage.

%% file: table_openwam_weak_tasks.tex
hanging\_mug & 43/54 & 36/34 & -13.5\\
turn\_switch & 72/65 & 45/57 & -17.5\\
open\_microwave & 96/73 & 73/71 & -12.5\\
blocks\_ranking\_size & 78/77 & 76/83 & +2.0\\
press\_stapler & 90/94 & 79/91 & -7.0\\
stack\_bowls\_three & 85/80 & 80/82 & -1.5\\
pick\_diverse\_bottles & 83/84 & 81/88 & +1.0\\
place\_can\_basket & 78/76 & 81/79 & +3.0\\
move\_stapler\_pad & 98/98 & 83/91 & -11.0\\
put\_object\_cabinet & 88/92 & 85/75 & -10.0\\

%% file: table_lila_training_prefix.tex
lila\_stage1 & 12 & 57.35 & 0.0689 \\
lila\_stage2 & 4 & 20.59 & 0.0641 \\

%% file: table_lila_weak_tasks.tex
place\_a2b\_left & 3/5 & 1/5 & -40 \\
place\_dual\_shoes & 4/5 & 1/5 & -60 \\
move\_stapler\_pad & 3/5 & 2/5 & -20 \\
open\_microwave & 4/5 & 2/5 & -40 \\
place\_mouse\_pad & 4/5 & 3/5 & -20 \\
click\_bell & 5/5 & 4/5 & -20 \\
hanging\_mug & 3/5 & 4/5 & +20 \\
lift\_pot & 5/5 & 4/5 & -20 \\
pick\_diverse\_bottles & 3/5 & 4/5 & +20 \\
place\_a2b\_right & 5/5 & 4/5 & -20 \\

%% file: sections/assessment.tex
\section{The Proposed RoboSeed Evaluation Framework}\label{sec:assessment}
\subsection{Decision Unit, Eligibility, and Interpretation}
We now formalize the framework introduced in Section 1 and apply it to the evidence collected in Chapters~\ref{sec:vla} and~\ref{sec:wam}. Our goal is to give community researchers a reusable assessment of task performance, resource feasibility and experimental support. The five dimensions answer complementary questions; the preference settings and eligibility rules connect them to a specific research decision.

A scientifically useful assessment must distinguish \emph{what a checkpoint can do}, \emph{what it costs}, and \emph{what the experiment establishes about its mechanism}. A single success leaderboard conflates these questions. We assess every reproduced policy family and every available RoboTwin checkpoint, including incomplete candidates. The unit is a \textbf{checkpoint--training recipe--interface--evaluator tuple}; the PX harness is a systems intervention on an existing unit. Broader ABot results supply transfer evidence and diagnostic context, but LIBERO percentages, RoboCasa percentages and CALVIN chain lengths are not averaged into RoboTwin capability.

Five dimensions, each on a 0--100 scale, define the profile: capability $P$, downstream training affordability $T$, deployment affordability $D$, locally verified research value $V$, and reproduction evidence $R$. Larger values are preferable under the stated rules. This rubric is a \textbf{versioned decision policy}, not an externally validated universal measure of algorithm quality. Thresholds and weights express explicit budget preferences; experimental counts and measured costs supply the empirical inputs. We state both, retain the raw metrics, and publish sensitivity to alternative preferences.

Three comparability gates precede a choice. First, the task interface must meet the application requirement: VTT task selection and free-form language are different capabilities. Second, training scope must be consistent: downstream adaptation of a released foundation does not include its inherited pre-train cost. Third, a causal architecture ranking requires matched datasets, input/actuation interfaces, precision, execution semantics, hardware and multiple training seeds. Our cross-method matrix does not pass the third gate. Scores can guide follow-up and budget screening; cross-method differences cannot be interpreted as isolated architecture effects.

\subsection{Quantitative Scoring Rules}
\subsubsection{Capability and Sampling Uncertainty}
For the shared RoboTwin target, let $p_C$ and $p_R$ be task macro-average success fractions in Clean and Randomized. Define
\begin{equation}
 P=100\,(p_C+p_R)/2.\label{eq:p-score}
\end{equation}
All observed scenes have equal denominators per task, so the macro- and micro-averages coincide there. Both scene results are required to compute $P$. If either scene is missing, $P$ is shown as a dash (--) rather than a numerical range or an imputed value. LingBot official and LiLa official therefore have no two-scene $P$ score, while their observed \emph{Clean-only} values remain 89.72\% and 89.60\% in the method-specific result tables. A completed training run without rollout outcomes also receives a dash for $P$.

Episode sampling uncertainty is calculated only from observed outcomes. Machine-readable output includes a working Wilson 95\% interval for each observed scene; for $s$ successes in $n$ episodes, with $\widehat p=s/n$ and $z=1.96$, it uses
\begin{equation}
 \frac{\widehat p+z^2/(2n)\ \pm\ z\sqrt{\widehat p(1-\widehat p)/n+z^2/(4n^2)}}{1+z^2/n}.
\end{equation}
Pooling is only a working binomial approximation for a fixed tested task mixture. Task heterogeneity, shared seeds, simulator nondeterminism and training variation can violate its assumptions. We do not use overlapping intervals to prove equivalence, or small point differences to establish superiority. No multi-seed training uncertainty is estimated from one checkpoint.

\subsubsection{Training Affordability and Lifecycle Scope}
For a positive cost $x$, use a fixed-anchor log utility
\begin{equation}
 F(x;a,b)=100\,\operatorname{clip}_{[0,1]}\!\left(\frac{\log(b/x)}{\log(b/a)}\right),\quad 0<a<b.\label{eq:utility}
\end{equation}
A cost at or below $a$ earns 100, a cost at or above $b$ earns zero, and multiplicative changes have equal utility increments between these anchors. Anchors remain fixed when another method is added, avoiding sample-dependent min--max rankings. A zero score means exceeding a chosen budget anchor, not scientific worthlessness.

Training score is
\begin{equation}
 H=\sum_j N_j t_j,\qquad T=F(H;64,1024),\label{eq:t-score}
\end{equation}
where $N_j$ is the allocated H200 GPU count and $t_j$ is the recorded training interval in hours. The anchors represent 16 hours and 256 hours on four GPUs; within this range each doubling costs 25 score points. All \emph{scored} local intervals are on H200 hardware. An RTX-5090 hour is not substituted for an H200 hour. Official checkpoints with no measured reference training runtime receive a dash for $T$, even when reference updates and batch are known.

$T$ measures \emph{the observed downstream recipe's device cost}, not compute to a common target success rate. The stopped LingBot run can therefore have a favorable $T$ and a poor $P$ simultaneously; this is why $T$ must never be read alone. It is not time-to-90\%, unique-data efficiency, FLOPs, energy, utilization or price. Approximate legacy training memory is retained in Table~\ref{tab:train}, but is not used in $T$ because sampling conventions and units are not uniform. A candidate that exceeds an actual GPU's memory capacity is infeasible regardless of its score.

For a deployment project, a more complete cost is
\begin{equation}
 C_{\mathrm{project}}=C_{\mathrm{inherited\ pretrain}}^{\mathrm{allocated}}+
 C_{\mathrm{adapt}}+C_{\mathrm{search/failures}}+K C_{\mathrm{evaluation}}+
 C_{\mathrm{serving}}+C_{\mathrm{integration}}.
\end{equation}
Only selected terms are measured here. In particular, foundation pre-train is inherited for LingBot and OpenWAM, VLM/vision priors are inherited by ABot and LiLa, failed trials are not comprehensively metered, and engineering person-hours are unavailable. Hence neither a dollar-denominated total cost nor an end-to-end training-cost winner is reported.

\subsubsection{Deployment Affordability and Required Measurements}
Deployment has distinct storage, memory and responsiveness costs. Define
\begin{equation}
 D=0.2F(W;2,32)+0.4F(M;8,80)+0.4F(L_{95};50,1000),\label{eq:d-score}
\end{equation}
where $W$ is the primary released model-weight artifact size in decimal GB, $M$ is warmed policy inference peak memory in GiB, and $L_{95}$ is p95 policy-request latency in milliseconds. The respective anchors encode a 2--32 GB distribution budget, 8--80 GiB serving budget and 50--1000 ms request budget. These are declared assessment preferences, not robotics safety deadlines. The primary artifact term measures distribution burden only; separately required encoders, teachers or caches must also be inventoried before procurement. It does not establish the complete installation footprint.

A valid latency measurement must record the same accelerator, precision, batch one, warmup and request count, camera inputs, executed/predicted horizon, integration steps and whether preprocessing/transfer/decoding are included. Latency should include those policy-request components while excluding simulator reset and physics. Sampling several hundred warmed requests with raw timings is required before scoring $M$ and $L_{95}$; simulator throughput, a mean request time or a nominal parameter count is insufficient. A robot-specific requirement must additionally compare latency to its control deadline and open-loop action horizon.

No policy in this snapshot has the required paired memory and p95 trace, so every $D$ entry is a dash. Only the primary-artifact component is available for ABot, LingBot and OpenWAM; its measured bytes and utility are retained in the raw metadata and subscores, without presenting that partial component as the full $D$ score. LiLa's checkpoint packaging is not comparable. Training VRAM is not substituted for inference VRAM, and a frozen encoder still incurs inference cost. A small parameter count motivates a future measurement but does not supply a measured latency score.

\subsubsection{Locally Verified Algorithmic Value}
Research value is evaluated through the strength of \emph{locally established, reusable conclusions}, rather than novelty adjectives or paper reputation. Four yes/no criteria each contribute 25 points:
\begin{enumerate}
\item \textbf{Executable mechanism:} an inspected implementation and effective configuration identify the proposed mechanism and its conditioning/actuation path.
\item \textbf{Decision comparator:} local outcomes include a documented within-family reference or controlled alternative that supports a concrete decision, with confounds disclosed. An unrelated published percentage is insufficient.
\item \textbf{Mechanism isolation:} a local component ablation holds data, exposure, interfaces and evaluator fixed and is repeated over at least three training seeds. A whole-checkpoint comparison does not qualify.
\item \textbf{Transfer test:} the same local checkpoint is tested without adaptation on a documented new distribution, with outcomes and denominators. Independently retrained checkpoints on several benchmarks do not alone qualify.
\end{enumerate}
Thus $V=25\sum_{k=1}^4 v_k$. A criterion is zero when qualifying local evidence has not been supplied; this evaluates the evidence portfolio, not an unknown future capability. Original-paper ablations motivate hypotheses but are not counted as locally reproduced evidence. The score is assigned at the method-family level and shared across its checkpoints. It must not be taken to imply that every individual backbone has a transfer result.

\begin{table}[!htbp]\centering\small
\caption{Research-value decisions and their local evidence. A one denotes a satisfied criterion. No local mechanism-isolation experiment meets the stated rule.}\label{tab:value}
\begin{tabularx}{\linewidth}{C{2.3cm}C{1.0cm}C{1.0cm}C{1.0cm}C{1.0cm}Y}\toprule
Family & Impl. & Comp. & Isolate & Xfer. & Evidence and $V$\\\midrule
ABot-M0 & 1 & 1 & 0 & 1 & Backbone budget comparison; same LIBERO checkpoint on LIBERO-plus. $V=75$.\\\addlinespace[3pt]
LingBot 2.0 & 1 & 1 & 0 & 0 & Official/local recipe gap, common local evaluator; no isolated MoE/teacher effect. $V=50$.\\\addlinespace[3pt]
OpenWAM & 1 & 1 & 0 & 0 & Official/local adaptation gap on the same execution path; randomized data used in training. $V=50$.\\\addlinespace[3pt]
LiLa-WAM & 1 & 1 & 0 & 0 & VTT inspected; official/local stage-2 comparison on a documented common evaluator. $V=50$.\\\bottomrule
\end{tabularx}\end{table}

These scores explain \emph{which claims the reproduction supports}. ABot's extra points reflect the local comparison and robustness portfolio; they do not prove AML is intrinsically more valuable than predictive world modeling. LingBot and OpenWAM provide testable predictive mechanisms, while LiLa offers a compact, language-free route with a different application boundary. Quantifying their individual contributions requires the missing ablations.

\subsubsection{Reproduction Evidence}
Five binary checks each contribute 20 points to $R$: (i) a configuration and weight-to-run mapping exists; (ii) original task records support the reported counts; (iii) every observed scene covers all 50 tasks with at least 20 episodes each; (iv) the local training interval and allocated device count are documented; and (v) at least three independent training seeds have complete evaluations. These are audit milestones, not a probability that a result is correct. Task totals reconstructed from aggregate rates remain weaker than independent episode traces even when check (ii) passes. Episode-level completeness and seed manifests should accompany future audits.

Evaluated local ABot, LingBot and OpenWAM units earn 80; the two latter official-checkpoint units earn 60 because reference training runtime is unavailable. Both LiLa units earn 40: all tasks are covered, but five episodes do not pass (iii), and independently verified hardware-normalized training intervals are unavailable. The newly completed ABot Qwen3.5-2B unit earns 80 like the other evaluated ABot units. None satisfies the multi-seed check. Missing a Randomized scene affects $P$ and score coverage rather than silently failing the per-observed-scene audit rule.

\subsection{Aggregation, Missing Data, and Sensitivity}
For dimension vector $\mathbf{s}=(P,T,D,V,R)$, the balanced preference is
\begin{equation}
 S_{\mathrm{bal}}=0.30P+0.20T+0.20D+0.15V+0.15R.\label{eq:total-score}
\end{equation}
\textbf{Missing measurements are shown as dashes (--).} A composite dimension is calculated only when all its required components are available; the overall score is calculated only when every positively weighted dimension is available. Missing values are stored as \texttt{null} in JSON, rather than zero or a numerical range. Missing weights are not redistributed to observed terms. Thus all current aggregate scores are dashes because deployment measurements are incomplete. A measured utility of zero and an audited criterion earning zero points remain valid numbers; they are distinct from an unavailable measurement.

Weighted measurement coverage records available inputs without estimating the missing scores. Let $c_P$ be half the number of observed scene rates, $c_T$ indicate an available training-cost measurement, and $c_D=0.2I_W+0.4I_M+0.4I_L$, where each indicator is one when that deployment input is available. The evidence-audit dimensions have $c_V=c_R=1$, including audited criteria that earn zero points. For profile weights $w_j$, define $q=\sum_j w_j c_j$. This is the observed fraction of aggregate input weight, not a success probability or statistical confidence. Even when some inputs of a composite dimension are available, that dimension remains a dash until complete. An overall numerical score requires $q=100\%$; any ranking must additionally meet the interface/protocol gates and account for episode and training uncertainty. No overall numerical ranking is reported in this snapshot.

\begin{table}[!htbp]\centering\scriptsize
\caption{Complete assessment of the 12 checkpoint units. Q3-VL and Q3.5 abbreviate the backbones. $P$ is two-scene capability, $T$ training affordability, $D$ deployment affordability, $V$ research value, and $R$ reproduction evidence. A dash denotes unavailable measurements or an incomplete composite. $q$ reports available input weight. Measured scores are rounded only for display.}\label{tab:scores}
\setlength{\tabcolsep}{3pt}
\begin{tabularx}{\linewidth}{C{3.65cm}YYYYYYY}\toprule
Checkpoint & $P$ & $T$ & $D$ & $V$ & $R$ & $S_{\mathrm{bal}}$ & $q$\\\midrule
\input{table_assessment_rows.tex} 
\bottomrule\end{tabularx}\end{table}

For a worked example, ABot Qwen3-VL-2B has $P=83.78$, $H=214.049$ GPU-hours, $V=75$ and $R=80$, with $T=56.46$. Its 7.766 GB artifact supplies only the storage component of $D$; inference memory and p95 latency are unavailable. Therefore both $D$ and the overall score are dashes. Available weighted inputs give $q=0.30+0.20+0.20\times0.20+0.15+0.15=84\%$. This describes evidence coverage and does not assign an overall performance score. The calculation is reproducible from \texttt{assessment.json}.

\fig{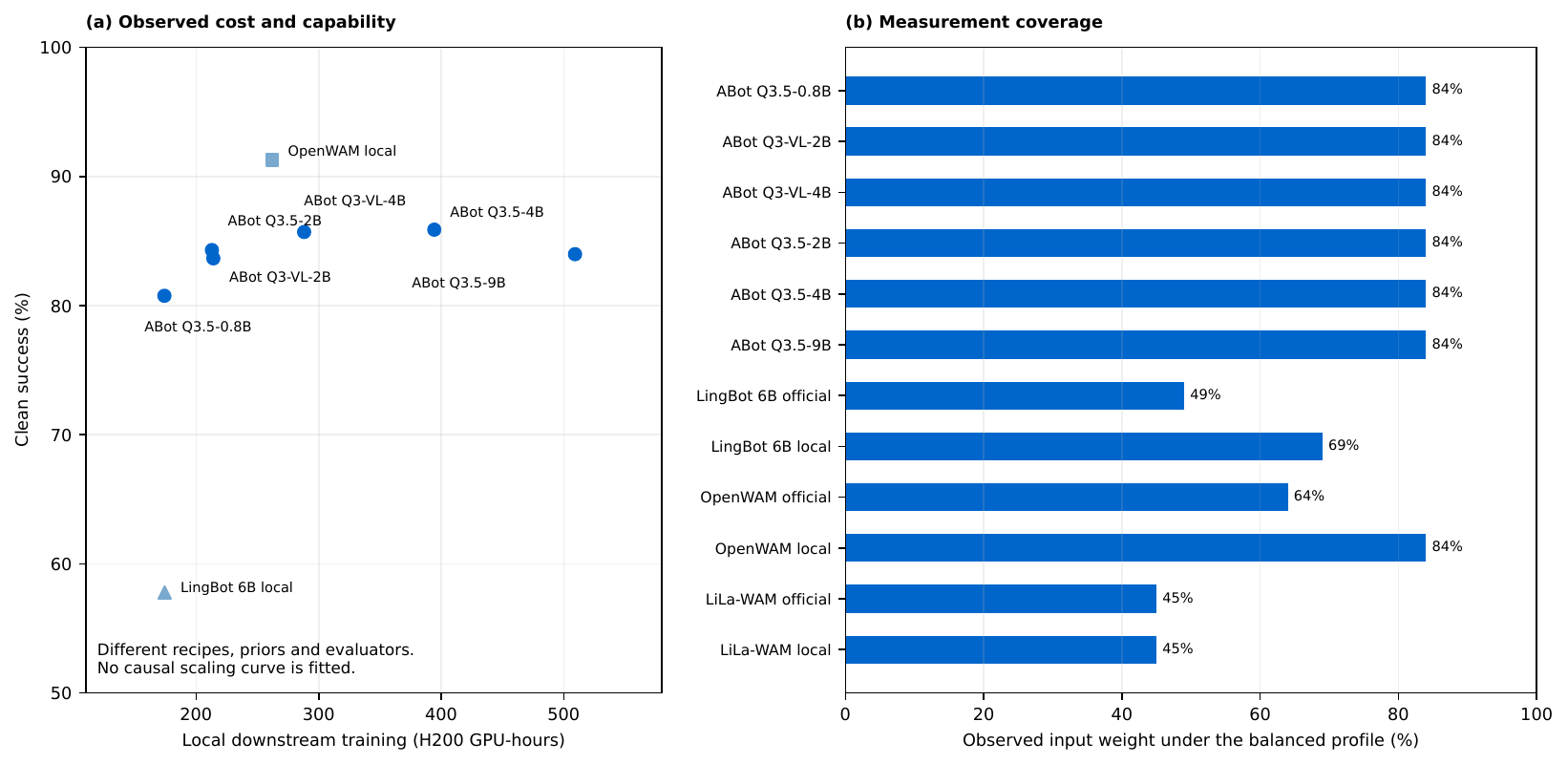}{1}{Cost--capability observations and measurement coverage. (a) Only local checkpoints with measured downstream H200 intervals and observed Clean outcomes are plotted. Different initialization, interfaces and protocols prevent a causal scaling interpretation. (b) Bars show available input weight under the balanced profile for all 12 units. Coverage is not an aggregate model score; incomplete scores are shown as dashes in the tables.}{fig:assessment}

\begin{table}[!htbp]\centering\small
\caption{Preference sensitivity. Weights follow the order $(P,T,D,V,R)$ and sum to one. These are explicit decision preferences rather than empirically learned constants.}\label{tab:weights}
\begin{tabularx}{\linewidth}{C{3.8cm}YY}\toprule
Profile & Weights & Use case\\\midrule
Balanced & $(.30,.20,.20,.15,.15)$ & General reproduction decision\\
Training constrained & $(.30,.35,.10,.10,.15)$ & Repeated adaptation with limited compute\\
Deployment constrained & $(.30,.10,.35,.10,.15)$ & Serving memory and latency dominate\\\bottomrule
\end{tabularx}\end{table}

\begin{table}[!htbp]\centering\scriptsize
\caption{Aggregate availability and weighted measurement coverage under each preference profile. Each cell is aggregate score / $q$. The aggregate remains a dash until all required inputs are available.}\label{tab:sensitivity}
\begin{tabularx}{\linewidth}{C{5.0cm}YYY}\toprule
Checkpoint & Balanced & Training constrained & Deployment constrained\\\midrule
\input{table_assessment_sensitivity.tex} 
\bottomrule\end{tabularx}\end{table}

Changing preferences changes input coverage, but cannot supply missing serving measurements. The deployment-constrained profile assigns more weight to the missing deployment inputs, reducing $q$ for most units. All three profiles therefore withhold the overall score. Once the inputs are complete, their aggregate scores can be compared under the stated weights. Threshold sensitivity should also be assessed if a target budget differs from our anchors; the released scoring function makes those preferences editable. Neither the balanced weights nor the anchor choices are fitted to reproduce a desired method ranking.

\subsection{Scientific Conclusions and Practical Choices}
\textbf{The completed 2B result refines the observed budget frontier.} Among the matched ABot runs, 0.8B remains the lowest-cost option at 80.63\% and 174.1 H200 GPU-hours. Qwen3.5-2B reaches 84.12\% at 212.9 GPU-hours, and Qwen3-VL-4B reaches 85.52\% at 288.1 GPU-hours. Relative to Qwen3-VL-2B (83.78\%, 214.0 hours), the Qwen3.5-2B point estimate is slightly better at slightly lower cost; the 0.34-point difference requires multi-seed validation. Both 4B variants exceed the 9B run's observed success while using fewer GPU-hours. These observations define a recipe-level point-estimate frontier, not a seed-robust architecture ranking.

\textbf{A capability floor should precede cost optimization.} An illustrative 200-H200-GPU-hour adaptation budget and an observed two-scene success floor of 80\% admits ABot 0.8B. A 250-hour budget and 84\% floor admits the newly completed Qwen3.5-2B candidate; a 300-hour budget and 85\% floor admits Qwen3-VL-4B. Qwen3-VL-2B also remains feasible under a 250-hour, 83\% requirement. These are screens using observed checkpoints, not guarantees for a new training seed or an untested task mixture.

\textbf{OpenWAM is a strong candidate for a harmonized follow-up.} It combines 91.35\% observed two-scene success with 262.068 recorded downstream GPU-hours, but deployment traces and a matched ABot evaluator are missing. Its strong recipe-level result justifies testing the same seed/instruction manifest and common control interface, followed by joint video/action ablations. It does not yet establish that the world model is the source of the gain or that total lifecycle cost is lower.

\textbf{LingBot exposes a budget-fidelity problem.} Its local recorded cost is similar to ABot 0.8B, but Clean success is much lower. Official weights perform substantially better on the same local evaluator. This combination favors investigating the adaptation recipe and exposure before scaling the architecture. Forecasts of 80--90\% after continuation in source notes are hypotheses and are not used as measured capability or as a basis for cost-to-target claims.

\textbf{LiLa tests a different service contract.} Its compact fixed-task solution is attractive when tasks can be selected through precomputed visual conditions. Current Clean success supports a larger-sample follow-up, not a language-conditioned ranking or a verified consumer-GPU deployment claim. The local stage-2 rollout now provides 222/250 (88.80\%) Clean successes, compared with 224/250 (89.60\%) for official weights. The next decisive evidence is larger matched sampling, Randomized coverage, a complete DINO-plus-action inference bundle, and direct memory/latency measurements. For applications requiring arbitrary language instructions, the present interface does not satisfy the capability gate regardless of its numerical Clean rate.

\textbf{Deployment engineering can create value independently of policy learning.} PX demonstrates about $2\times$ end-to-end throughput improvement for similar same-concurrency workloads without retraining. The result is actionable for simulation-scale evaluation. Its seed/instruction differences and mean inference statistics do not meet the serving p95 protocol or establish unchanged accuracy on identical episodes. An engineering comparison should therefore report workload integrity, end-to-end time, device-hours, actual steps and policy latency separately.

The highest-value next measurements are those that narrow a decision-relevant gap: serving traces for all policies, Randomized coverage for LingBot and LiLa, and higher-coverage LiLa outcomes. For mechanism research, repeat controlled ablations across seeds. For operational selection, first enforce language/task interfaces, memory and deadlines, then compare feasible profiles under declared budget weights. This turns the reproduction matrix into a set of falsifiable decisions rather than a catalog of method names.

%% file: table_assessment_rows.tex
ABot Q3.5-0.8B & 80.6 & 63.9 & -- & 75.0 & 80.0 & -- & 84\% \\
ABot Q3-VL-2B & 83.8 & 56.5 & -- & 75.0 & 80.0 & -- & 84\% \\
ABot Q3-VL-4B & 85.5 & 45.7 & -- & 75.0 & 80.0 & -- & 84\% \\
ABot Q3.5-2B & 84.1 & 56.6 & -- & 75.0 & 80.0 & -- & 84\% \\
ABot Q3.5-4B & 85.5 & 34.4 & -- & 75.0 & 80.0 & -- & 84\% \\
ABot Q3.5-9B & 83.8 & 25.2 & -- & 75.0 & 80.0 & -- & 84\% \\
LingBot 6B official & -- & -- & -- & 50.0 & 60.0 & -- & 49\% \\
LingBot 6B local & -- & 63.9 & -- & 50.0 & 80.0 & -- & 69\% \\
OpenWAM official & 93.8 & -- & -- & 50.0 & 60.0 & -- & 64\% \\
OpenWAM local & 91.3 & 49.2 & -- & 50.0 & 80.0 & -- & 84\% \\
LiLa-WAM official & -- & -- & -- & 50.0 & 40.0 & -- & 45\% \\
LiLa-WAM local & -- & -- & -- & 50.0 & 40.0 & -- & 45\% \\

%% file: table_assessment_sensitivity.tex
ABot Q3.5-0.8B & -- / 84\% & -- / 92\% & -- / 72\%\\
ABot Q3-VL-2B & -- / 84\% & -- / 92\% & -- / 72\%\\
ABot Q3-VL-4B & -- / 84\% & -- / 92\% & -- / 72\%\\
ABot Q3.5-2B & -- / 84\% & -- / 92\% & -- / 72\%\\
ABot Q3.5-4B & -- / 84\% & -- / 92\% & -- / 72\%\\
ABot Q3.5-9B & -- / 84\% & -- / 92\% & -- / 72\%\\
LingBot 6B official & -- / 49\% & -- / 42\% & -- / 47\%\\
LingBot 6B local & -- / 69\% & -- / 77\% & -- / 57\%\\
OpenWAM official & -- / 64\% & -- / 57\% & -- / 62\%\\
OpenWAM local & -- / 84\% & -- / 92\% & -- / 72\%\\
LiLa-WAM official & -- / 45\% & -- / 40\% & -- / 40\%\\
LiLa-WAM local & -- / 45\% & -- / 40\% & -- / 40\%\\

%% file: sections/px_harness.tex
\subsection{PX Evaluation Harness: Systems Reproduction}\label{sec:harness}
\subsubsection{What Is Being Compared}
The fourth source is a deployment/evaluation study of \textbf{the existing ABot/PX-M0 Qwen3-VL-2B policy}, not a fourth new action algorithm. It uses three-camera input, 14 absolute joint channels, 50-action chunks, BF16, eager attention and ten DDIM steps. These settings differ from the original ABot action-integration configuration. The October 28 refresh adds complete 50-task systems workloads alongside the original subsets. Historical 100-episode capability scores remain distinct from these ten-episode systems tests.

The original entry launches one model per task process. The harness uses one model service per GPU, simulation containers that share an episode queue, reusable expert-validated seeds and reduced planner initialization for joint control. The implementation can batch requests, but cache access and actual model inference are distinguished. This changes the resource-allocation strategy and makes environment/model dependencies easier to maintain independently.

\subsubsection{Experimental Setup}
\begin{table}[!htbp]\centering\small
\caption{PX harness systems-study settings. These experiments benchmark an existing VLA policy rather than training another action model.}\label{tab:harness_setup}
\begin{tabularx}{\linewidth}{C{4.2cm}Y}\toprule
Setting & Observed configuration\\\midrule
Policy / accelerator & ABot/PX-M0 Qwen3-VL-2B / H200 NVL\\
Inputs / action interface & Three cameras / 14 absolute joint channels\\
Chunk / numerical path & 50 actions / BF16, eager attention, ten DDIM steps\\
Single-GPU timed workload & 12 Randomized tasks $\times$ ten episodes\\
Original / shared model instances & One per task process / one per GPU\\
Primary concurrency / batch cap & Six- and eight-worker comparisons / one\\
Startup / rollout videos & Included in end-to-end timing / videos disabled\\
Cross-framework inputs & Accepted seeds and instruction RNG differ; actual steps reported\\\bottomrule
\end{tabularx}\end{table}

\subsubsection{Measured Speed, Bottlenecks, and Limits}
\begin{table}[!htbp]\centering\small
\caption{Single-GPU systems comparison on 12 Randomized tasks, ten episodes each. End-to-end time includes model startup. Historical PX time is reconstructed from task-directory and log timestamps. Seed acceptance and instruction sampling differ across frameworks, so this is an approximate workload comparison.}\label{tab:harness}
\begin{tabularx}{\linewidth}{C{3.8cm}YYYY}\toprule
Entry & Workers & Successes & Executed steps & End-to-end seconds\\\midrule
Original PX & 6 & 98/120 & 39,692 & 2418.949\\
Harness W6/B1 & 6 & 101/120 & 39,352 & 1182.580\\
Harness W8/B1 & 8 & 98/120 & 41,175 & 1100.330\\\bottomrule
\end{tabularx}\end{table}

At equal six-worker concurrency, end-to-end speedup is $2418.949/1182.580=2.045\times$; executed-step throughput improves from 16.41 to 33.28 steps/s ($2.028\times$). W8 supplies a further $1.075\times$ wall-time speedup. These gains support the combined system design; they do not isolate model sharing, initialization, queue scheduling or seed reuse individually. No VRAM trace was collected, so a measured peak-memory saving cannot be claimed.

In a separate same-subset scaling run, two GPUs with eight workers each complete 120 episodes in 619.499 seconds, versus 1002.641 seconds on one GPU. Wall time improves by $1.62\times$, but allocation rises from 0.279 to 0.344 GPU-hours (about 23.6\% more). More GPUs can minimize completion time while increasing device cost. The throughput of 11.62 episodes/minute is specific to that subset and cannot be extrapolated directly to all 50 tasks.

\begin{table}[!htbp]\centering\small
\caption{Harness GPU-allocation experiments on the separate 120-episode subset. Evaluation-only times exclude startup and must not be mixed with Table~\ref{tab:harness}'s end-to-end times.}\label{tab:harness_scaling}
\begin{tabularx}{\linewidth}{C{4.1cm}YYYY}\toprule
Deployment & GPUs & Wall seconds & GPU-hours & Episodes/min\\\midrule
Co-located W8/B1 & 1 & 1002.641 & 0.279 & 7.18\\
Co-located W8/B1 per GPU & 2 & 619.499 & 0.344 & 11.62\\
Split model/simulation W16/B1 & 2 & 814.538 & 0.453 & 8.84\\\bottomrule
\end{tabularx}\end{table}

Profiling explains why increasing batch limits does not reliably help. In the split-service B4 retry, 97.87\% of 7,969 action requests read cached actions; 170 genuine inference requests produce 147 model calls, mean batch 1.156, with 89.1\% of calls still at batch one. Around 95\% of cumulative episode time is reset, rendering, observation and physics; the remaining 5\% is a residual including networking and waiting, not pure model computation. In that workload, even eliminating the entire residual would yield only about $1/0.95=1.053\times$ speedup if all other costs remained fixed. This is a workload-specific Amdahl bound, not a universal limit for VLA inference. Co-location increases the residual and contention.

The harness uses prevalidated seed lists and an independent instruction RNG. A source audit found different accepted seed lists for 11 of 12 sampled tasks relative to the historical entry. Thus 98/120 versus 101/120 cannot establish a capability improvement or strict correctness equivalence. The safe conclusion is an approximately twofold systems speedup for similar workloads, with success counts, actual steps, startup scope and protocol differences disclosed. The harness is assessed as an engineering intervention rather than assigned a separate policy capability score.

\subsubsection{Full-Task Validation and Updated Throughput Comparison}
The October 26 runs extend the comparison to all 50 Randomized tasks with ten episodes each on two GPUs. All three entries in Table~\ref{tab:harness_full} complete 500 episodes and exit successfully. Original PX task results match their final ten-episode log counters; harness aggregate task files independently reproduce their complete, issue-free summaries. Startup-inclusive timings come from the launchers, while rollout-only times remain separate.

\begin{table}[!htbp]\centering\small
\caption{Updated two-GPU systems comparison over all 50 Randomized tasks. W denotes workers per GPU; harness batch cap is one. Both GPUs are allocated throughout each end-to-end interval. Seed acceptance and instruction sampling differ across execution paths.}\label{tab:harness_full}
\setlength{\tabcolsep}{3pt}
\begin{tabularx}{\linewidth}{C{3.4cm}YYYYY}\toprule
Entry & Successes & Steps & End-to-end (s) & GPU-hours & Episodes/min\\\midrule
Original PX / W6 & 411/500 & 136,145 & 4185.99 & 2.326 & 7.17\\
Harness / W6 & 408/500 & 138,487 & 2222.52 & 1.235 & 13.50\\
Harness / W16 & 401/500 & 145,542 & 1814.54 & 1.008 & 16.53\\\bottomrule
\end{tabularx}\end{table}

At matched W6 concurrency, the full-task harness run has $1.88\times$ higher episode throughput and $1.92\times$ higher executed-step throughput than the original PX entry. W16 reaches $2.31\times$ and $2.47\times$, respectively, relative to original PX. The W16 versus W6 end-to-end speedup is $1.225\times$; actual rollout step counts increase by about 5.1\%. Rollout-only times are 2177.461 and 1690.594 seconds for W6/W16, excluding the 45.059/123.946-second startup difference. These results support throughput gains over the full tested task set, while protocol and realized-workload differences preclude attributing the success-count differences to the model or interpreting the speed ratios as isolated component effects.

The refreshed seed-fallback test completes 42/50 successes over five tasks in 354.304 rollout seconds. Six episodes use replacement seeds. In the full-task W6 and W16 runs, 32 and 24 episodes use replacements, respectively. Seed manifests therefore remain necessary even when task names and episode counts match. Additional one-GPU 12-task runs complete 99/120 successes at W12 in 973.547 seconds and 95/120 at W16 in 954.563 seconds, both measured without startup. Their small timing difference is an observed workload result, not a demonstrated universal worker optimum.

Two earlier four-GPU full-task runs record 500 rows but explicitly fail completeness checks: W6 reports infrastructure/partial-result issues on 17 tasks, and W16 on 11 tasks. Their success-rate fields are null, so they are retained as failure diagnostics rather than promoted to capability or throughput winners. A separate Qwen3.5-2B harness launch exits with code one after 7.90 seconds and has no completed summary; this does not invalidate that checkpoint's independently completed historical 100-episode evaluation. The accessible-run inventory, source hashes and task-level system records are retained in \path{data/current_results_audit.json} and the refresh evidence directory.

\subsubsection{Four-GPU Topology and Batch-Cap Comparison}
After access was restored, the October 28 batch-comparison directory supplied three further complete, issue-free 120-episode evaluations. All aggregate files reproduce the summary counts. The 12 task names, requested/actual seeds and instructions match across the three runs. Centralized deployment reserves one GPU for the policy server and three for simulation, with eight workers per simulation GPU (24 total); co-location uses four policy services and 16 workers on each of four GPUs (64 total). The worker counts therefore differ as well as the topology.

\begin{table}[!htbp]\centering\small
\caption{Four-GPU topology and batch-cap observations on the matched 120-episode input manifest. B is the configured maximum model batch, not a measured mean batch size. Rollout-only and startup-inclusive times are separated.}\label{tab:harness_oct28}
\setlength{\tabcolsep}{3pt}
\begin{tabularx}{\linewidth}{C{3.1cm}YYYYYY}\toprule
Deployment & Workers & B & Successes & Steps & Rollout (s) & End-to-end (s)\\\midrule
Centralized (1+3 GPUs) & 24 & 1 & 101/120 & 37,918 & 405.435 & 485.22\\
Centralized (1+3 GPUs) & 24 & 4 & 95/120 & 42,856 & 472.496 & 529.27\\
Co-located (4 GPUs) & 64 & 1 & 99/120 & 41,997 & 414.647 & 471.01\\\bottomrule
\end{tabularx}\end{table}

Centralized B1 has the shortest rollout-only interval, while co-located B1 has the shortest startup-inclusive interval. Increasing the centralized batch cap to four does not reduce the observed wall time, but that run also executes about 13.0\% more steps than centralized B1. The approximately 16.5\% rollout-time increase cannot therefore be attributed entirely to batching overhead. The cap alone does not establish that inference actually formed batches of four. Although the episode inputs match, differing realized success counts and step lengths, single runs, and the changed simulation concurrency limit causal interpretation. These observations support measuring both startup and realized workload when choosing an evaluation topology; they do not establish a capability ranking or a universal optimal batch size.

%% file: sections/extension_appendix.tex
\section{New Reproduction Counts and Provenance}\label{app:extension}
The October 28 refresh rechecks the supplied experiment roots against the previous report. It adds the completed ABot Qwen3.5-2B and local LiLa-WAM outcomes, updates their resource/evidence records, and incorporates the latest accessible PX systems runs. Existing ABot backbone, LIBERO, LIBERO-plus, RoboCasa and CALVIN observations remain unchanged where no new outcome is found. The extension manifest now lists 235 source records, with byte counts and SHA-256 hashes; a separate refresh audit inventories new, incomplete and inaccessible candidates. Large weights are neither loaded nor copied nor hashed: artifact sizes, saved states, configurations and resolved checkpoint paths identify the studied weights, without claiming a cryptographic tensor audit.

\texttt{scripts/collect\_data.py} now validates all six ABot RoboTwin configurations against their final 100-episode task counters. \texttt{scripts/collect\_extension.py} imports those 600 task/scene rows and adds 400 LingBot/OpenWAM/LiLa rows, yielding 1,000 unique checkpoint/scene/task records. LiLa official and local worker JSONs are independently checked against their final log counters and corresponding checkpoint settings by \texttt{scripts/audit\_current\_results.py}. The official backup directory is kept separate from the reused current result directory. OpenWAM's existing rate records retain the documented 100-episode denominator; no independent episode-verdict reconstruction is claimed for them.

The refresh audit also checks every accessible harness summary and retains unsuccessful runs. Two new full-task harness runs reproduce all 500 episode outcomes from 50 aggregate task files each; the original PX full-task comparator reproduces 411/500 successes and 136,145 executed steps from task logs. Startup-inclusive launcher intervals remain distinct from rollout-only summary times. The Wan2.1-VACE-1.3B saved 50k state has no rollout results and remains outside the evaluated checkpoint profiles. Access to the previously restricted October 28 four-GPU batch-comparison directory was restored. All three 120-episode runs are now checked against their task aggregates and included in Section~\ref{sec:harness}; their task/seed/instruction manifests match. Exact scanned paths, access limits, configurations and evidence hashes are retained in \path{data/current_results_audit.json}.

The offline \path{scripts/assessment.py} regenerates \path{data/assessment.json}, the score and sensitivity rows, the new result/task tables and Figure~\ref{fig:assessment}. Its inputs are the saved unit metadata in \path{data/extension_units.json} and task counts in \path{data/extension_tasks.csv}, with policy version 1.1. The JSON preserves per-scene Wilson intervals, all value/evidence flags, raw costs, scoring anchors, preference weights, scalar scores and input coverage. Missing measurements and incomplete composites are \texttt{null} in JSON and dashes in tables. Version 1.1 replaces the earlier missing-input ranges with this explicit missing-value policy; observed measurements and scoring anchors are unchanged. New measurements under the same rules should update the evidence snapshot. Source paths and hashes are in \path{data/extension_manifest.json}, and copied records are under \path{data/extension_sources/}.

\subsection{Per-Method Experimental Results and Training Curves}
Chapters~\ref{sec:vla} and \ref{sec:wam} place experimental results immediately after each method description. The additional training observations are generated by \path{scripts/method_experiments.py} with source access using \texttt{--collect}, or regenerated offline from the saved CSVs without that flag. The separate \path{data/method_experiments_manifest.json} records six source hashes, extraction decisions, native first/last losses, and LiLa epoch timers. The original checkpoint outcomes and scoring rules remain unchanged by this extraction.

LingBot contributes 50,000 update records; OpenWAM contributes 80,000 selected progress-counter records with four-decimal display precision and a possible one-update offset. LiLa contributes 23,643 and 7,881 sampled-update records for completed stage-1 and stage-2 prefixes. Its seven/four partial-next-epoch observations are excluded explicitly. All retained observations remain in the CSVs even when light figure traces are subsampled; moving averages use the full series. Model-native loss components are plotted separately and are not averaged across methods. The logging implementations used to interpret OpenWAM counters and LiLa loss components are copied into the method-experiment source directory.

\subsection{All 50 New Task Counts}
L-O and L-L denote official and local LingBot Clean successes out of 100. W-O and W-L give official and local OpenWAM Clean/Randomized successes, each out of 100. Li-O and Li-L give official and local LiLa Clean counts out of five. LiLa and LingBot Randomized remain unavailable in this evidence set. Sections~\ref{sec:lingbot}, \ref{sec:harness}, \ref{sec:openwam} and \ref{sec:lila} describe the different protocols; row alignment does not make episodes paired.

\begingroup\scriptsize
\setlength{\tabcolsep}{4pt}
\begin{longtable}{lrrrrrr}
\caption{Complete task counts for the new model reproductions.}\label{tab:new_tasks}\\
\toprule Task & L-O & L-L & W-O (C/R) & W-L (C/R) & Li-O & Li-L\\\midrule\endfirsthead
\toprule Task & L-O & L-L & W-O (C/R) & W-L (C/R) & Li-O & Li-L\\\midrule\endhead
\midrule\multicolumn{7}{r}{Continued on next page}\\\endfoot
\bottomrule\endlastfoot
\input{table_new_tasks.tex} 
\end{longtable}\endgroup

%% file: table_new_tasks.tex
adjust\_bottle & 100 & 99 & 100/99 & 96/99 & 5/5 & 5/5\\
beat\_block\_hammer & 91 & 45 & 98/99 & 100/97 & 4/5 & 5/5\\
blocks\_ranking\_rgb & 92 & 58 & 100/96 & 98/100 & 5/5 & 5/5\\
blocks\_ranking\_size & 76 & 36 & 78/77 & 76/83 & 4/5 & 5/5\\
click\_alarmclock & 64 & 62 & 97/100 & 96/99 & 5/5 & 5/5\\
click\_bell & 58 & 18 & 100/100 & 100/100 & 5/5 & 4/5\\
dump\_bin\_bigbin & 96 & 90 & 97/97 & 96/97 & 4/5 & 5/5\\
grab\_roller & 100 & 100 & 100/99 & 100/99 & 5/5 & 5/5\\
handover\_block & 95 & 5 & 96/97 & 98/89 & 4/5 & 5/5\\
handover\_mic & 93 & 91 & 89/95 & 94/96 & 5/5 & 5/5\\
hanging\_mug & 48 & 44 & 43/54 & 36/34 & 3/5 & 4/5\\
lift\_pot & 100 & 16 & 100/100 & 100/99 & 5/5 & 4/5\\
move\_can\_pot & 88 & 16 & 99/99 & 96/98 & 5/5 & 5/5\\
move\_pillbottle\_pad & 95 & 57 & 99/99 & 100/96 & 5/5 & 5/5\\
move\_playingcard\_away & 100 & 91 & 100/100 & 100/99 & 5/5 & 5/5\\
move\_stapler\_pad & 70 & 33 & 98/98 & 83/91 & 3/5 & 2/5\\
open\_laptop & 98 & 97 & 99/99 & 96/100 & 5/5 & 5/5\\
open\_microwave & 84 & 62 & 96/73 & 73/71 & 4/5 & 2/5\\
pick\_diverse\_bottles & 85 & 14 & 83/84 & 81/88 & 3/5 & 4/5\\
pick\_dual\_bottles & 89 & 10 & 95/98 & 94/88 & 5/5 & 5/5\\
place\_a2b\_left & 93 & 82 & 97/98 & 98/95 & 3/5 & 1/5\\
place\_a2b\_right & 94 & 91 & 100/96 & 98/92 & 5/5 & 4/5\\
place\_bread\_basket & 93 & 62 & 94/97 & 96/90 & 4/5 & 4/5\\
place\_bread\_skillet & 94 & 59 & 90/88 & 88/89 & 4/5 & 4/5\\
place\_burger\_fries & 96 & 73 & 97/99 & 99/96 & 5/5 & 5/5\\
place\_can\_basket & 63 & 38 & 78/76 & 81/79 & 3/5 & 4/5\\
place\_cans\_plasticbox & 89 & 2 & 99/100 & 98/95 & 5/5 & 5/5\\
place\_container\_plate & 96 & 95 & 100/97 & 100/99 & 5/5 & 5/5\\
place\_dual\_shoes & 93 & 49 & 96/91 & 93/87 & 4/5 & 1/5\\
place\_empty\_cup & 100 & 61 & 100/100 & 100/100 & 5/5 & 4/5\\
place\_fan & 84 & 58 & 97/95 & 99/100 & 5/5 & 5/5\\
place\_mouse\_pad & 97 & 51 & 99/99 & 97/98 & 4/5 & 3/5\\
place\_object\_basket & 92 & 56 & 93/89 & 90/86 & 3/5 & 5/5\\
place\_object\_scale & 95 & 78 & 95/99 & 88/93 & 5/5 & 5/5\\
place\_object\_stand & 96 & 69 & 97/97 & 99/93 & 5/5 & 5/5\\
place\_phone\_stand & 90 & 76 & 93/95 & 94/99 & 5/5 & 5/5\\
place\_shoe & 99 & 84 & 100/100 & 99/98 & 5/5 & 5/5\\
press\_stapler & 100 & 93 & 90/94 & 79/91 & 5/5 & 5/5\\
put\_bottles\_dustbin & 83 & 13 & 94/95 & 89/91 & 5/5 & 5/5\\
put\_object\_cabinet & 96 & 62 & 88/92 & 85/75 & 5/5 & 5/5\\
rotate\_qrcode & 94 & 73 & 93/96 & 91/89 & 4/5 & 4/5\\
scan\_object & 96 & 16 & 97/94 & 95/89 & 5/5 & 5/5\\
shake\_bottle & 100 & 96 & 100/100 & 99/100 & 5/5 & 5/5\\
shake\_bottle\_horizontally & 100 & 99 & 100/100 & 99/100 & 5/5 & 5/5\\
stack\_blocks\_three & 97 & 36 & 97/96 & 91/93 & 5/5 & 5/5\\
stack\_blocks\_two & 100 & 46 & 100/100 & 100/100 & 5/5 & 5/5\\
stack\_bowls\_three & 78 & 45 & 85/80 & 80/82 & 5/5 & 4/5\\
stack\_bowls\_two & 96 & 82 & 97/94 & 95/96 & 5/5 & 5/5\\
stamp\_seal & 78 & 37 & 96/98 & 87/95 & 2/5 & 5/5\\
turn\_switch & 82 & 63 & 72/65 & 45/57 & 4/5 & 4/5\\